\documentclass{article}
\usepackage{ijcai25}

\usepackage{times}
\usepackage{soul}
\usepackage{url}
\usepackage[hidelinks]{hyperref}
\usepackage[utf8]{inputenc}
\usepackage[small]{caption}
\usepackage{graphicx}
\usepackage{amsmath}
\usepackage{amsthm}
\usepackage{booktabs}
\usepackage{algorithm}
\usepackage{algorithmic}
\usepackage[switch]{lineno}
\usepackage{multirow}
\usepackage{adjustbox}
\usepackage{tabularx}
\usepackage{amsfonts}
\usepackage{subcaption}
\usepackage[normalem]{ulem}
\usepackage{caption}
\usepackage{float} 
\usepackage{subcaption}
\usepackage{siunitx}
\usepackage{appendix}
\usepackage[table,xcdraw]{xcolor}
\newtheorem{thm}{Remark}[section]

\title{HeTa: Relation-wise Heterogeneous Graph Foundation Attack Model}

\author{
Yuling Wang$^{1,2}$
\and
Zihui Chen$^{1,2}$\and
Pengfei Jiao$^{1,2}$\footnote{Corresponding author.}\And
Xiao Wang$^3$\\
\affiliations
$^1$ School of Cyberspace, Hangzhou Dianzi University\\
$^2$ Data Security Governance Zhejiang Engineering
Research Center, Hangzhou Dianzi University
$^3$Beihang University
\\
\emails
\{wangyl0612, 231270002, pjiao\}@hdu.edu.cn,
xiao\_wang@buaa.edu.cn
}

\begin{document}

\maketitle

\begin{abstract}
Heterogeneous Graph Neural Networks (HGNNs) are vulnerable, highlighting the need for tailored attacks to assess their robustness and ensure security.
However, existing HGNN attacks often require complex retraining of parameters to generate specific perturbations for new scenarios.
Recently, foundation models have opened new horizons for the generalization of graph neural networks by capturing shared semantics across various graph distributions.
This leads us to ask:
\textit{Can we design a foundation attack model for HGNNs that enables generalizable perturbations across different HGNNs, and quickly adapts to new heterogeneous graphs (HGs)?}
Empirical findings reveal that, despite significant differences in model design and parameter space, different HGNNs surprisingly share common vulnerability patterns from a relation-aware perspective. 
Therefore, we explore how to design foundation HGNN  attack criteria by mining shared attack units.  
In this paper,  we propose a novel relation-wise heterogeneous graph foundation attack model, HeTa. 
We introduce a foundation surrogate model to align
heterogeneity and identify the importance of shared relation-aware attack units. 
Building on this, we implement a serialized relation-by-relation attack based on the identified relational weights. 
In this way, the perturbation can be transferred to various target HGNNs and easily fine-tuned for new HGs. Extensive experiments exhibit powerful attack performances and generalizability of our method. 


\end{abstract}

\section{Introduction}
Heterogeneous graphs (HGs), prevalent in fields like  biological networks \cite{ma2023single} and knowledge graphs \cite{san2024kg}, realistically represent complex systems with diverse nodes and edges across different domains.
Existing heterogeneous graph neural networks (HGNNs) are often vulnerable to adversarial attacks due to perturbation amplification from the complex coupling of heterogeneous nodes and relations  \cite{zhang2022robust}.
Therefore, developing tailored attack methods for HGNNs is essential for assessing their robustness and ensuring the security of their applications. 

Current attacks on HGs primarily utilize gradients from surrogate models with structures similar to the target model to perform topology attacks, thereby degrading the performance of the target HGNN.
Broadly speaking, these attacks mainly fall into two groups: (1) targeted attack, which aim to reduce the performance of specific target instances but non-target might remain unchanged to avoid being detected \cite{wang2024unsupervised}; and (2) global attacks, which have recently emerged to reduce the overall performance of HGNNs with limited budget \cite{shang2023transferable}. 
While promising, existing methods on HGs rarely account for the generalization of attackers, which necessitates retraining attack parameters for different target HGNNs and new HGs. 
Although some recent works have attempted transferable attacks \cite{shang2023transferable,zhao2024hgattack}, they fail to achieve cross-domain applicability and require elevated permissions to manipulate either the graph structure or the training data.

Due to their powerful generalization capabilities, foundation models have revolutionized the fields of Natural Language Processing~\cite{qin2023chatgpt} and Computer Vision~\cite{liu2024sora}, demonstrating significant potential in learning general, open-world knowledge from diverse data sources. This has endowed them with strong expressiveness and adaptability, enabling them to excel across a wide range of tasks and datasets. Building on this concept, Graph Foundation Models (GFMs) constitute a significant advancement in graph-structured data, facilitating cross-domain and cross-task generalization through the training of a unified, transferable vocabulary (e.g., basic graph elements like relations and subgraphs). This advancement prompts us to consider a natural question: 
 \textit{Can we design a foundation attack model for HGNNs that enables generalizable perturbations across different HGNNs, and quickly adapts to new HGs?}

\begin{figure}
    \centering
    \includegraphics[width=0.55\linewidth]{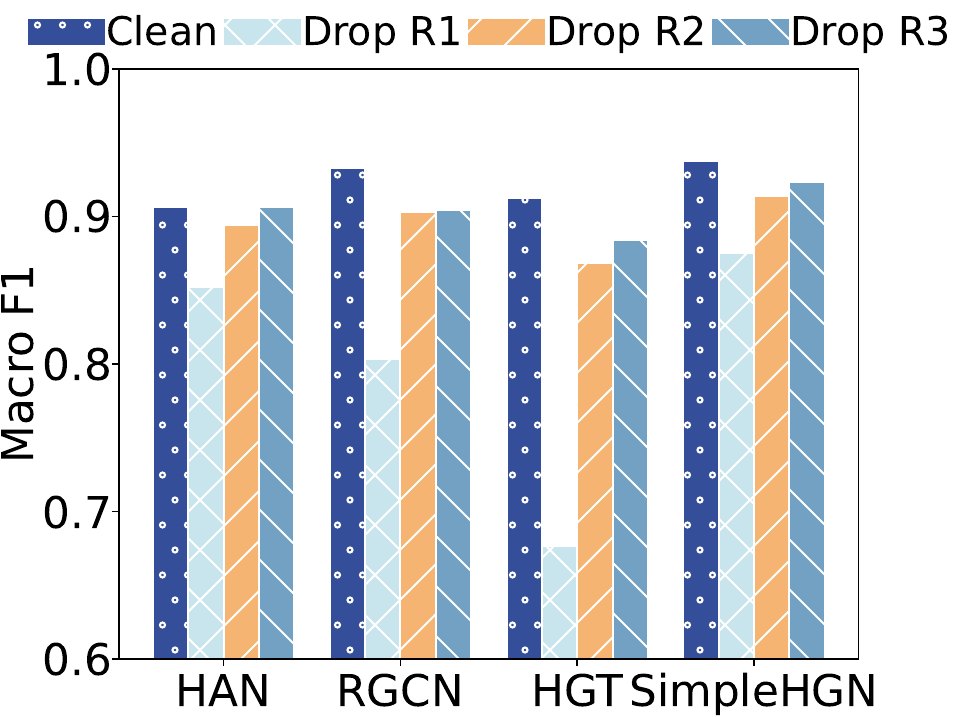}
    \caption{Performance of different HGNNs on the ACM dataset with various relations dropped, where `Clean' denotes the original graph (R1: author-paper,  R2: paper-subject, R3: paper-term).}
    \label{fig:motivate}
\end{figure}

To address this question, we empirically investigate the impact of removing different relation subgraphs from various HGNNs on the ACM dataset as Figure \ref{fig:motivate}, as the relationships among nodes often serve as the fundamental semantic of HGs \cite{wang2022ensemble,wang2019heterogeneous}. 
The results clearly show that the importance of the relations is consistent across these HGNNs, i.e., R1$>$R2$>$R3. Specifically, removing relation R1 typically leads to the most significant performance drop across all HGNNs (up to 23.6\%), while removing R3 causes only a minor impact (up to 2.8\%).  It indicates that these HGNNs share a common importance pattern across relational subgraphs and are likely to be consistently vulnerable when the most crucial relation is perturbed.  This motivates us to view relations as the fundamental shared attack unit,  laying the groundwork for a foundation attack model for HGNNs. 

Despite its potential, achieving this goal entails significant challenges.
\textbf{First}, 
how to identify the shared importance distribution of
attack semantic units across different HGNNs?
The differences in structural and feature distribution between  HGs are considerable~\cite{xia2024}. Therefore, different HGNNs requires careful heuristic design and generally involves significantly large and diverse parameter spaces to model semantic diversity \cite{wang2019heterogeneous}, making it difficult to uncover the underlying generalizable principles shared across HGNNs.
\textbf{Second}, once the importance pattern is captured, how to design universal attack criteria based on this pattern to progressively destroy each critical  semantic unit in the HG?
Since the HG is decomposed into different attack units with varying semantics, it is crucial to attack these units precisely, step by step, based on their importance, while employing a low-budget and low-permission strategy.




In this paper, we propose a  relation-wise \underline{He}terogeneous graph foundation at\underline{Ta}ck model (HeTa), which identifies shared attack patterns based on relational semantics, enabling the attack process on HGs to generalize.
Specifically, we develop a lightweight foundation surrogate model to provide a unified characterization of different HGNNs, i.e., simplifying their propagation mechanisms to model the importance distribution of shared relational semantic units. A simple parametric mechanism that enables fast generalization to new graphs.
Subsequently, we implement a relation-by-relation serialized attack process based on the learned relation weights. This involves injecting adversarial nodes into each relation subgraph, with gradients from the carefully designed attack loss guiding the generation of fake edges, enabling a low-budget attack that requires minimal permissions to perturb the original topology. 
Finally, the perturbed HG can be fed into different target HGNNs to evaluate the attack's effectiveness and generalizability, and the attack on a new HG can be quickly fine-tuned by freezing part of the attacker's parameters. Our key contributions are as follows:


\begin{itemize}
\item 
To our knowledge, it is the first foundation attack model for HGNNs. 
We also verify that different HGNNs share common vulnerabilities in relation-aware attack units.
\item 
We propose HeTa, a novel generalizable HGNN attack model, that enables transferability across different target HGNNs and easily adapts to new HGs.
\item Extensive experiments on three public datasets demonstrate the effectiveness and generalizability of our proposed HeTa under node injection and evasion attacks.
\end{itemize}
\section{Related Work}
\subsection{Heterogeneous Graph Neural Network}
HGNNs are widely acclaimed for enhancing node representations. According to the way to handle heterogeneity, HGNNs mainly fell into meta-path-aware and relation-aware methods. 
The former relies on carefully designed meta-paths to aggregate information, often involving complex relation combinations.~\cite{fu2024mecch,wang2019heterogeneous}. 
The latter is aggregate messages from neighbors  of different relations \cite{yu2022multiplex,yang2023simple}.
Recent advancement have focused on leveraging LLM to achieve generalization across diverse  HGs~\cite{tang2024higpt}). 
AnyGraph learns a foundation model from a vast of graphs to effectively handle the heterogeneity \cite{xia2024}. Despite significant differences, these HGNNs essentially regard relations as the most fundamental semantic units.



\subsection{Adversarial Attack on Graph Neural Network}
Graph adversarial attacks have  gained traction because minimal perturbations can mislead models. Homogeneous graph attack, such as \cite{chen2018fast} and \cite{goodfellow2014explaining} use gradient information to guide attack edges. \cite{zhang2024maximizing} proposes a susceptible-reverse influence sampling strategy for selecting neighbors. While in HG, the adversarial robustness remains less explored. Roughly speaking,  attacks in HGNNs fell into target attack and global attack. The former manipulate graph to mislead target instances,
e.g., \cite{zhao2024hgattack} proposes a semantic-aware mechanism to automatically identify and generate  perturbations to mislead target nodes. While the latter aim to destroy the total performance, e.g., \cite{shang2023transferable} proposes a structured global attack method guided by edge attention. While promising, no existing study has yet explored a generalizable foundation model for HGNNs.

\section{Preliminaries}
\subsection{Heterogeneous Graph}
Heterogeneous Graph (HG), \(\mathcal{G} = (\mathcal{V}, \mathcal{E},\mathcal{F})\), consists of an node set \(\mathcal{V}\), an edge set \(\mathcal{E}\) and a feature set \(\mathcal{F}\). 
\(\mathcal{G}\) is also associated with a node type mapping function \(\phi: \mathcal{V} \rightarrow \mathcal{T}\) and an edge type mapping function \(\psi: \mathcal{E} \rightarrow \mathcal{R}\). \(\mathcal{T}\) and \(\mathcal{R}\) denote the predefined type sets of node and edge, where \(|\mathcal{T}| + |\mathcal{R}| > 2\). Let \( V_{\tau} \) denotes the node set of type \( \tau \in \mathcal{T} \), the feature set \( \mathcal{F} \) is composed of \( |\mathcal{T}| \) feature matrices, \( \mathcal{F} = \{F_{\tau}, \tau \in \mathcal{T}\} \), \( F_{\tau} \in \mathbb{R}^{|V_{\tau}| \times d_{\tau}} \), where \( d_{\tau} \) is the feature dimension of \( \tau \) nodes.
\paragraph{Relation Subgraph.} Given a HG, \(\mathcal{G} = (\mathcal{V}, \mathcal{E},\mathcal{F})\),  \( \mathcal{G}_r \) is a subgraph of \( \mathcal{G}\) that contains all edges of relation \( r \). The adjacency matrix of \( \mathcal{G}_r \) is \( A_r \in \mathbb{R}^{N \times N} \), 
where $N$ is the number of nodes in $\mathcal{G}$.
\( A_r[i, j] = 1 \) if \( \langle v_i, r, v_j \rangle \) exists in \( \mathcal{G}_r \), otherwise \( A_r[i, j] = 0 \),
where $ v_i, v_j \in \mathcal{V}$, $r \in \mathcal{R}$. \( \mathcal{A} \) is adjacency matrix for $\mathcal{G} $, i.e.,
 \ \( \mathcal{A} = \sum_{r=1}^\mathcal{R} A_r\).

 \subsection{Node Injection Attack in HG}
Given a HG,  \(\mathcal{G} = (\mathcal{A},\mathcal{F})\), node injection attack (NIA) generates a fake node set \(N_{in}\) and connects to existing nodes in $\mathcal{G}$. The perturbed HG after injecting \(N_{in}\) can be denoted as \(\mathcal{G'} = (A'_{1}, A'_{2}, ..., A'_{|\mathcal{R}|}; F'_{1},F'_{2},...,F'_{|\mathcal{T}|})\). E.g., for two types of nodes in \(\mathcal{G}\), Author (A) and Paper (P), connected by edges type P-A, after injecting fake nodes $N_{in}$ of type P, we have:
\begin{align}
    A'_{PA} = \begin{bmatrix}
            A_{PA} & E_{in} \\
            E_{in}^\top & E_{0}
        \end{bmatrix},
        F'_{P} = \begin{bmatrix}
            F_{P} \\
             F_{in}
        \end{bmatrix},
\end{align}%
where \( A_{PA} \in \mathbb{R}^{N \times N} \) is the subgraph of  P-A type, \( E_{in} \in \mathbb{R}^{N \times |N_{in}|} \) is the adjacency between the original and injected nodes, and \( E_{0} \in \mathbb{R}^{|N_{in}| \times |N_{in}|} \) is the adjacency between injected nodes. \( F_P \in \mathbb{R}^{N \times d_P} \)  and \( F_{in} \in \mathbb{R}^{ |N_{in}| \times d_P} \)  are the features of the original and the injected nodes, respectively.
\paragraph{Foundation Attacker’s Goal.}
The foundation attacker aims to construct generalizable perturbation graphs that degrade the target model’s performance, which can transfer to new scenarios, like new target models and graph domains.

\paragraph{Attacker’s Knowledge and Capability.} We focus on NIA in black-box and evasion settings, where the attacker can only modify the test data without access to the target model’s parameters or architecture \cite{sun2022adversarial}. 
Given a clean HG, \(\mathcal{G} = (\mathcal{A},\mathcal{F})\), we train a surrogate model \(f_\theta\) and freeze its parameters.
Next, the attacker is trained based on the surrogate model's behavior to generate fake nodes and inject them into $\mathcal{G}$, forming a perturbed graph, \(\mathcal{G'}=(\mathcal{A'},\mathcal{F'})\), within budget. Finally, we apply the attacked $\mathcal{G'}$ to other target models to reduce their predictions. The unified formulation is:
\begin{align}
& \max_{G'} \mathcal{L}\left(f_{\theta^*}\left(\mathcal{G'}\right)\right), 
\nonumber \\
& \text{s.t.} \quad \theta^* = \arg\min_{\theta}\mathcal{L}\left(f_{\theta}\left(\mathcal{G}\right)\right),  \label{eq:NIA_formulation} \\
& |N_{in}| \leq N*\rho, \quad \rm{deg}(v)_{v \in N_{in}} \leq K, \nonumber 
\end{align}%
where \(\theta^*\)  denotes the surrogate model's optimal parameters trained by the loss  \(\mathcal{L}\), i.e., the cross-entropy loss for node classification. The higher $\mathcal{L}\left(f_{\theta^*}\left(\mathcal{G'}\right)\right)$, the better the performance of the attack. The injected node set \(N_{in}\) is limited by a injected rate \(\rho\) and total node number \(N\) in $\mathcal{G}$, the degree of each injected node $v$ is limited by average degree \(K\) of $\mathcal{G}$.

\begin{figure*}[!ht]
    \centering
    \includegraphics[width=0.9 \textwidth]{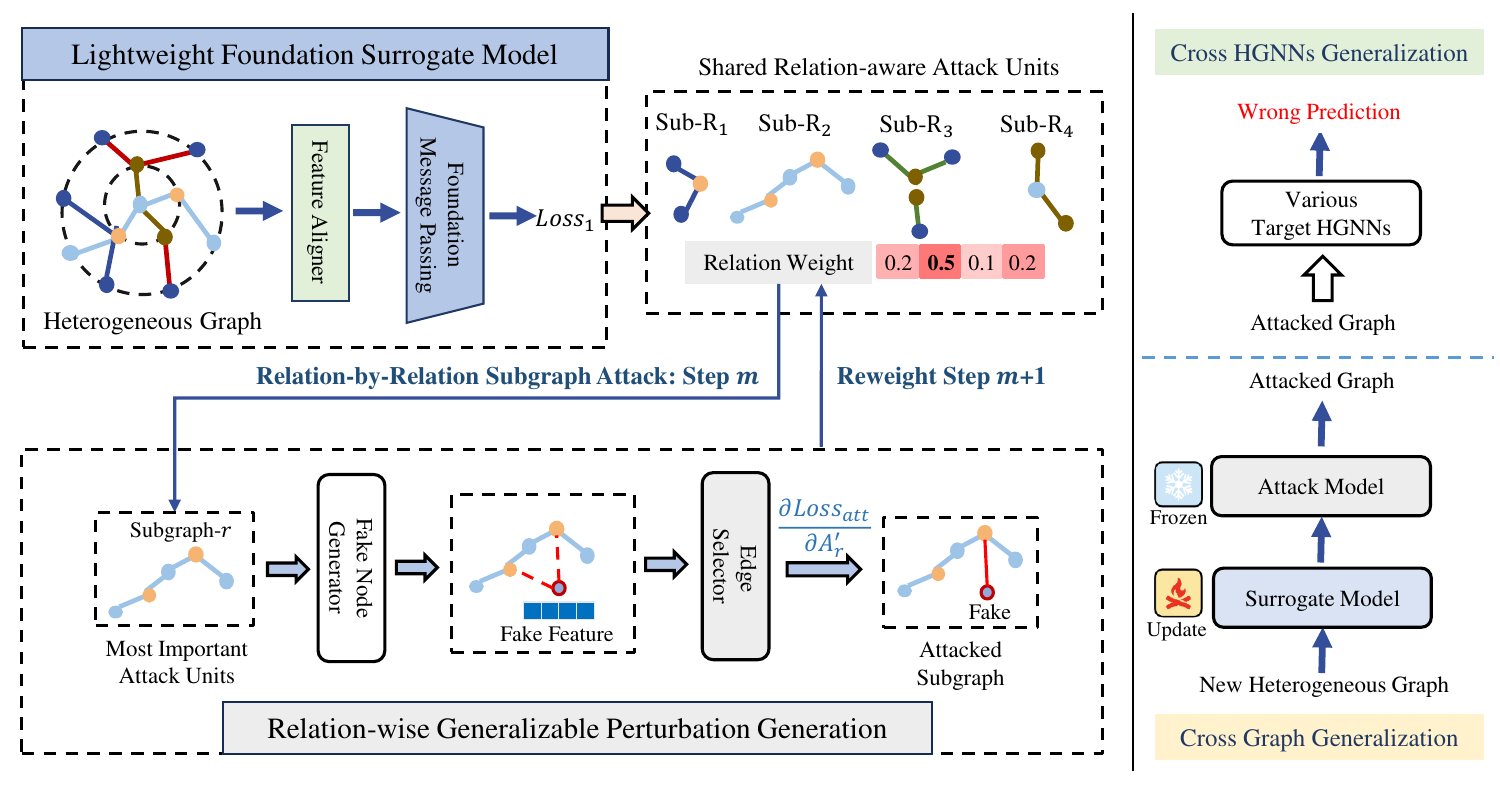}
    \caption{
An overview of the framework for our proposed HeTa model.
    }
    \label{fig:attack_process}
\end{figure*}

\section{Proposed Framework}
In this section, we introduce HeTa, a novel foundation attack model designed specifically for HGNNs by identifying relation-aware shared attack units. An overview of the framework is shown in Figure~\ref{fig:attack_process}.



\subsection{A Lightweight Foundation Surrogate Model }


We introduce a foundation surrogate model to represent the vulnerabilities of different HGNNs, aiming to achieve two key objectives:  (1) Generalization, providing a unified description of the common characteristics shared across various HGNNs,  and (2) Vulnerability Learning, identifying the importance distribution of fundamental attack units in HGs.

\paragraph{Heterogeneous Features Alignment.} 

Since different node types \(\tau\in\mathcal{T}\) in HGs typically come from inconsistent distributions, we first align the original features  \(h_v \in \mathbb{R}^{1 \times d_{\tau}}\)   of various node types into a unified vector space:
%
\begin{align}
     h_v^0 =\text{Projector}(h_v),\label{eq:projector}
\end{align}%
where $h_v^0$  denotes the unified node representations in a shared semantic space.
The feature aligner, i.e., $\text{Projector}(\cdot)$, can be flexibly configured as any trainable model for different node types; here, we instantiate it using a linear transformation.


\paragraph{Heterogeneous Foundation Message Passing.}

Building on the experimental results shown in Figure \ref{fig:motivate}, which reveal a unified vulnerability pattern across different HGNNs from a relational perspective, we propose treating relational subgraphs as the fundamental attack units. 
To this end, we employ an ensemble multi-relational message passing mechanism~\cite{wang2022ensemble} to model relational semantics in HGs using learnable coefficients, i.e., the weights of fundamental attack units in our case. Specifically, messages from different relation subgraphs are aggregated by weighted summation based on the learned relational coefficients:
\begin{align}
    H^{l} = \sigma \left[ \left(  \sum_{r=1}^\mathcal{R} \mu_r \hat{A}_r \right) H^{l-1} \right],
\end{align}%
where $\mu_{r}$  and  \(\hat{A}_r\) represent the weight and the normalized adjacency matrix of relation $r$, respectively,  with $\sum_{r=1}^\mathcal{R} \mu_r=1$. Given by \(\hat{A}_r = \tilde{D}_r^{-\frac{1}{2}} \tilde{A}_r \tilde{D}_r^{-\frac{1}{2}}\), \(\tilde{A}_r = A_r + I\) and \(\tilde{D}_r = D_r + I\), where \(D_r\) is the degree matrix of $A_r$.  \(\sigma\) denotes activation function, we use ReLU here.
To preserve the node's original information as much as possible and prevent over-smoothing, we employ residual connections:
\begin{align}
   Y_{\rm{pred}} = \text{Classifier}\left(  (1 - \alpha) H^l + \alpha H^0   \right),\label{eq:Z_softmax}
\end{align}%
where $H^0 =[h_1^0,h_2^0\dots,h_N^0]$  is the initial features from $\text{Projector}(\cdot)$,  and \(\alpha\in[0, 1]\) is a adjustable scaling factor. The $\text{Classifier}(\cdot)$ is used to output the probability distribution $Y_{\rm{pred}}$ of node labels and is implemented as an MLP in our work. We use cross-entropy loss for node classification as the training objective, as shown in Eq.\eqref{eq:NIA_formulation}.
Each element in \(\boldsymbol{\mu}=[\mu_{1}, \mu_{2}, \ldots, \mu_{|\mathcal{R}|}]\) is initially set to $ \frac{1}{|\mathcal{R}|}$, and adaptively learned during this training. 

Consequently,
the optimized $\boldsymbol{\mu}$ indicates the importance distribution of relational attack units, which identify common vulnerabilities across HGNNs. Furthermore, its lightweight parameterization enables rapid adaptation to new HGs, and we will validate its generalizability in the experiments.



\subsection{Relation-wise Generalizable  Perturbation Generation}
Traditional attacks on HGNNs typically generate perturbations based on an overall attack loss, without considering the disruption of each individual semantic component within the HG, leading to incomplete attacks. To alleviate this issue, 
 we leverage the importance distribution across various relation-aware attack units provided by the surrogate model and systematically attack each unit in a serialized manner—i.e., at each step, identifying and attacking the most important relation. In this way, the semantics of the entire HG will be progressively and thoroughly disrupted after $M$ steps.

\paragraph{Relation-wise Attack Principle.}
We identify fundamental attack units, relation weight $\boldsymbol{\mu}$ by surrogate model, and the relations with larger weights play a more crucial role in the model's training and prediction.
Based on this, we introduce a step-by-step relation-wise attack strategy to progressively perturb the most critical parts of the HG. The weights at step 0 are initialized as $\boldsymbol{\mu}$.
  Specifically, at step $m$, we attack the relation  unit $r$ with the largest weight:
\begin{align}
    r^{m} = \arg\max(\mu_{1}^{m}, \mu_{2}^{m}, \ldots, \mu_{|\mathcal{R}|}^{m}).\label{eq:relation_select}
\end{align}%
To prevent consecutive attacks from targeting the same relation, we introduce a penalty term \(\beta\)  to dynamic reweight the relation unit that was attacked in the previous step,  thereby obtaining the weights for next step: $ \mu_r^{m+1} = \frac{\mu_r^{m}}{\beta}$.


\paragraph{Attack Loss.}
During attacks, we reversely optimize the training objective of the surrogate model. Here, our loss function \(\mathcal{L}_{\rm{atk}}\)  includes the Carlini-Wagner (CW) attacks loss \cite{carlini2017towards} \(\mathcal{L}_{\rm{cw}}\) and the inverse of the KL divergence \cite{ji2020kullback} \(\mathcal{L}_{\rm{kl}}\). The objective of the attack is to minimize the \(\mathcal{L}_{\rm{atk}}\):
\begin{align}
    \mathcal{L}_{\rm{atk}} = \mathcal{L}_{\rm{cw}} + \mathcal{L}_{\rm{kl}}.\label{eq:loss_l}
\end{align}%
The CW loss minimizes the difference between the correct category and the maximum probability of other categories:
\begin{align}
     \mathcal{L}_{\rm{cw}} = \sum_{v \in V} \max \left\{\left( p_{v,c} - \max_{y_v \neq c} p_{v,y_v}\right), -k \right\}, \label{eq:loss_cw}
\end{align}%
where \(p_{v,c}\) denotes the probability that node \(v\) is predicted to be target class $c$ and \(\max_{y_v \neq c} p_{v,y_v}\) represents the probability of the most likely incorrect class. \(V\) is test node set and \(k\) (set as 0 here) is a confidential level of making wrong predictions.
The $ \mathcal{L}_{\rm{kl}}$ shifts the model's probability distribution:
\begin{align}
   -\mathcal{D}_{\rm{kl}}( \hat{y}_v \parallel y_{v})
   =\ln p_{v} ,
\end{align}%
where  \(\mathcal{D}_{\rm{kl}}\) is KL divergence, $\hat{y}_v$ the predicted label distribution from surrogate model for node \(v \in V\),
and $ y_{v}$ is the groundtruth.
\(p_v\) denotes  the predicted probability that $v$ belongs to its true category.
To prevent gradient explosion  when \(p_v\rightarrow0\), we use a smooth loss function \cite{zou2021tdgia}:
\begin{align}
   \mathcal{L}_{\rm{kl}} = \frac{1}{|V|} \sum_{v \in V} \max(r + \ln p_v, 0)^2, \label{eq:loss_v}
\end{align}%
where \(r\) is a control factor (set as 4 here).

\paragraph{Relation-wise Attack.}
At each attacking step $m \in [1, M]$, the disruption of the relation-aware semantic unit $r$ involves two phases:  first, generate the corresponding fake node set $N_{in}^{r}$, including their type and features, based on the relation of the current attack unit; second, create fake edge set $E_{in}^{r}$ connecting these fake nodes, guided by the gradient information of the current relation subgraph. 

\textit{(1) Fake Node Generator.} 
Based on the current attack relation $r$ selected in Eq.\eqref{eq:relation_select}, we randomly choose the injected node type from the head and tail entity types of relation 
$r$, denoted as \(\phi_{in}(r)\). 
The remaining unselected type is then designated as the connecting node type,  denotes as \(\phi_{con}(r)\). 
Here, we define the initial features and connections of the injected node $v_{in} \in N_{in}^{r}$. Specifically, we assume that $v_{in}$ is initially connected to all nodes of type \(\phi_{con}(r)\), and this connection will be further optimized by the edge selector. 
To enhance the imperceptibility of $v_{in}$,  we calculate the cluster center of all nodes of type \(\phi_{con}(r)\)
as the prototype $h_{con}$, then randomly sample a node feature $x_{in}^{0}$ close to $h_{con}$ as the initial feature of $v_{in}$.
To optimize the fake node's features for a more effective attack, we define a specific fake node generator \(f_g^{\phi_{in}(r)}\) for node type $\phi_{in}(r)$ as follows:
\begin{equation}
 x_{in}^{m} = f_g^{\phi_{in}(r)} \left( x_{in}^{m-1} + x_{\rm{neighbor}}^{m-1} \right), \label{node_generator}
\end{equation}
where $x_{in}^{m}$ is the feature of fake node $v_{in}$  at time step $m$,  with an  initial state  $x_{in}^{0}$.
$x_{\rm{neighbor}}^{m}$  corresponds to the feature aggregation of $v_{in}$'s neighbors. 
We set the fake node generators for different node types as distinct MLPs here.




\textit{(2) Fake Edge Selector.}
Then, the attacker will determine which target nodes in \(\mathcal{G}\) to connect with the injected $v_{in} \in N_{in}^{r}$, aiming to maximize disruption of the graph information.
A straightforward approach is to use the gradient of $\mathcal{L}_{\rm{cw}}$ with respect to the relation subgraph to guide the attack, targeting the positions where the gradient changes most significantly \cite{wang2020scalable}. However, applying back-propagation to compute the gradients of each fake node's connections is challenging for large-scale graphs.  We utilize an approximation strategy to simplify, after injecting \(v_{in}\), we linearize the surrogate model with multi-layer graph convolution and precompute the gradient \(\frac{\partial \mathcal{L}_{\rm{cw}}}{\partial A^{'m}_{r}}\) in forward propagation, then selecting top-K absolute value as neighbors for \(v_{in}\):
\begin{equation}
[A_r^{'m}]_{:, j}= \text{topK}\left\{\text{abs}([\frac{\partial \mathcal{L}_{\rm{cw}}}{\partial A_{r}^{'m-1}}]_{:, j})\right\},\label{fake_edge_selector}
\end{equation}
where  \(A_r^{'m}\) is the attacked  relation 
subgraph at the $m$-th  step after injecting \(v_{in}\),
and \([\cdot]_{:, j}\) is the \(j\text{-th}\)  column in the matrix, i.e., fake node. Details are in the Appendix A.

\subsection{Applications of Perturbation}
Built on the foundation principles of the surrogate model and relation-wise attack, the perturbed HG, $\mathcal{G'} = (\mathcal{A'}, \mathcal{F'})$, generalizes to degrade downstream HGNNs and quickly adapts to new HGs through simple fine-tuning.
\paragraph{Degrading Target HGNNs.} 
The attacked $\mathcal{G'}$ can be directly used to degrade predictions of various target models:
\begin{align}
  \hat{Y}_{\text{target}} = f_{\text{target}}\left(\mathcal{A'}, \mathcal{F'}\right),\label{eq:target_model_softmax}
\end{align}%
where $f_{\text{target}}(\cdot)$ can be any trained target HGNN, without the need to modify its internal architecture or parameters. $\hat{Y}_{\text{target}}$  is the prediction on the perturbed HG, which performance will significantly decrease compared to the original HG.

\paragraph{Adaptation to New HGs.} 
HeTa has two parts of trainable parameters, i.e., the surrogate model and the fake node generators.
 Assume we have trained a surrogate model $f_{\theta^*}$ and a set of fake node generators 
$f_{g^*} = \left\{f_{g^*}^{(k)}\right\}_{k=1}^K$ on  $\mathcal{G}_1$.
 Now, for a new graph  $\mathcal{G}_2$ that may have a different distribution  from $\mathcal{G}_1$, we can quickly adapt to $\mathcal{G}_2$ by freezing  $f_{g^*} $.
 Specifically, the $f_{\theta}$ can be easily retrained due to its lightweight nature. 
Then, we can randomly select $J$ frozen fake node generators from $f_{g^*}$ for the new graph $\mathcal{G}_2$, provided that $J <= K$.

\subsection{Additional Analysis}
\begin{thm}\label{theorem1}
Given a HG with a set of relation-aware attack units $\mathcal{A} = \{A_1,  A_2, ..., A_{|\mathcal{R}|}\}$, for each $A_r \in \mathcal{A}$, the vulnerability of the target node to the semantic unit $r$ increases as the degree of the node decreases.
\end{thm}
\paragraph{Proof:} We focus on the  attack unit $A_r$ when injecting $v_{in}$.
\begin{align}
    A'_r = \begin{bmatrix}
            A_r & e_r \\
            e_r^\top & 0
        \end{bmatrix},
    D'_r = \begin{bmatrix}
            D_r+d_1 & 0 \\
             0 & d_2
        \end{bmatrix},
\end{align}%
where $A'_r \in \mathbb{R}^{|N+1| \times |N+1|}$, 
$ e_r$ is the injected edge set, \(D'_r\) is the degree matrix of $A'_r$. \(d_1\) is the degree of the fake node with existing nodes, while \(d_2\) is the degree of the fake node (set as K). 
Given by \(\hat{A'}_r = \tilde{D}_r^{-\frac{1}{2}} (A'_r+I) \tilde{D}_r^{-\frac{1}{2}}\), \(\tilde{D}_r=D_r+I\).
For simplicity, let \(\lambda=\tilde{D}_r + d_1\), \(GD=\lambda^{-\frac{1}{2}} (A'_r+I) \lambda^{-\frac{1}{2}}\), \(d_r=d_2+1\). 
For the alignment feature $H$ from Eq.\eqref{eq:projector}, the learnable parameters in the surrogate model denotes as \(W_2\),
the gradient $\frac{\partial L_{cw}}{\partial e_{{r}}}$ can be derived as follows:
\begin{align}
   \alpha = 
    d_r^{-\frac{1}{2}} \lambda^{-\frac{1}{2}} \bigg[ d_r^{-\frac{1}{2}} H W_2 D_r^{*}  
    + [GD]_{:,\mathcal{I}_{V}} \ Diag(h_{in} W_2) \bigg],
\end{align}
where $ D_r^{*} $ represents the degree information of the  original graph $\mathcal{G}$ after injecting $v_{in}$.
$\mathcal{I}_{V}$ is index set for test nodes.
In this way, the first term can be ignored as it does not involve the injected fake nodes. We then analyze and simplify the second term as: 
\begin{equation}
\lambda^{-\frac{3}{2}}d_r^{-\frac{1}{2}} (A'_r+I).
\end{equation}
We observe that as  \(\lambda\) decreases, i.e.,  the degree of the target node decreases, the gradient increases, making the node more susceptible to attack. Details are provided in the Appendix A.



\section{Experiments}
\subsection{Experimental Settings}
\paragraph{Datasets.} In our experimental evaluation, we utilize three HG datasets: DBLP\footnote{\url{https://github.com/THUDM/HGB}\label{fn1:dataset}}, ACM\textsuperscript{\ref{fn1:dataset}}, and IMDB\footnote{\url{https://github.com/seongjunyun/Graph_Transformer_Networks}}. Details of these datasets are displayed in Appendix B.1.

\paragraph{Target HGNN Backbones.} We validate the effectiveness and generalizability of HeTa on four widely used HGNNs, i.e., HAN \cite{wang2019heterogeneous}, HGT \cite{hu2020heterogeneous}, RGCN \cite{schlichtkrull2018modeling}, SimpleHGN \cite{lv2021we}. Under evasion attacks, these HGNNs are trained on the clean graph and remain frozen parameters during evaluation.
\paragraph{Attack Baselines.} 
We compare two types of baselines: (1) Heterogeneous graph attacks: specifically \cite{zhang2022robust}, called RoHe-attack. Given the lack of injection attacks tailored for HGs, we also compare (2) Homogeneous graph attacks:  FGA \cite{chen2018fast}, \(\text{G}^2\text{A2C} \) \cite{ju2023let}.
The details are in the  Appendix B.2.

\paragraph{Implementation Details.} 
The implementation details of the experiments are provided in  Appendix B.3, and the pseudo-code for HeTa can be found in  Appendix C.



\begin{table*}
\centering
\resizebox{\linewidth}{!}{%
\begin{tabular}{c|c|c|cc|cc|cc|cc} 
\hline
\multirow{2}{*}{Dataset} & \multirow{2}{*}{Target Model} & \multirow{2}{*}{Attack Methods} & \multicolumn{2}{c|}{Clean}                        & \multicolumn{2}{c|}{1\%}          & \multicolumn{2}{c|}{2\%}          & \multicolumn{2}{c}{5\%}           \\ 
\cline{4-11}
                         &                               &                                 & Macro F1                & Micro F1                & Macro F1        & Micro F1        & Macro F1        & Micro F1        & Macro F1        & Micro F1         \\ 
\hline
\multirow{16}{*}{IMDB}   & \multirow{4}{*}{HAN}          & \(\text{G}^2\text{A2}\text{C} \)                           & \multirow{4}{*}{0.5206} & \multirow{4}{*}{0.5356} & 0.5200          & 0.5351          & 0.5190          & 0.5341          & 0.5140          & 0.5311           \\
                         &                               & FGA                             &                         &                         & 0.5157          & 0.5297          & 0.5093          & 0.5237          & 0.4859          & 0.5032           \\
                         &                               & RoHe-attack                     &                         &                         & 0.4950           & 0.5143          & 0.4672          & 0.4890           & 0.4583          & 0.4955           \\
                         &                               & HeTa                              &                         &                         & \cellcolor{gray!20} \textbf{0.3278} & \cellcolor{gray!20}\textbf{0.4106} & \cellcolor{gray!20}\textbf{0.2964} &\cellcolor{gray!20} \textbf{0.3578} &\cellcolor{gray!20} \textbf{0.2465} & \cellcolor{gray!20} \textbf{0.2879}  \\ 
\cline{2-11}
                         & \multirow{4}{*}{HGT}          & \(\text{G}^2\text{A2}\text{C} \)                           & \multirow{4}{*}{0.5487} & \multirow{4}{*}{0.5634} & 0.5424          & 0.5583          & 0.5336          & 0.5510          & 0.5033          & 0.5275           \\
                         &                               & FGA                             &                         &                         & 0.5336          & 0.5528          & 0.5219          & 0.5433          & 0.4838          & 0.5113           \\
                         &                               & RoHe-attack                     &                         &                         & 0.5423          & 0.5583          & 0.5346          & 0.5528          & 0.5053          & 0.528            \\
                         &                               & HeTa                              &                         &                         & \cellcolor{gray!20}\textbf{0.5133} & \cellcolor{gray!20}\textbf{0.5214} &\cellcolor{gray!20} \textbf{0.5049} & \cellcolor{gray!20}\textbf{0.5098} &\cellcolor{gray!20} \textbf{0.4624} & \cellcolor{gray!20}\textbf{0.4601}  \\ 
\cline{2-11}
                         & \multirow{4}{*}{SimpleHGN}    & \(\text{G}^2\text{A2}\text{C} \)                           & \multirow{4}{*}{0.5711} & \multirow{4}{*}{0.5874} & 0.5684          & 0.5848          & 0.5638          & 0.5801          & 0.5557          & 0.5720            \\
                         &                               & FGA                             &                         &                         & 0.5635          & 0.5788          & 0.5616          & 0.5775          & 0.5316          & 0.5408           \\
                         &                               & RoHe-attack                     &                         &                         & 0.5370          & 0.5540          & 0.4985          & 0.5151          & 0.3951          & 0.4023           \\
                         &                               & HeTa                              &                         &                         & \cellcolor{gray!20}\textbf{0.3436} & \cellcolor{gray!20}\textbf{0.4478} & \cellcolor{gray!20}\textbf{0.3274} & \cellcolor{gray!20}\textbf{0.4203} & \cellcolor{gray!20}\textbf{0.2825} & \cellcolor{gray!20}\textbf{0.3511}  \\ 
\cline{2-11}
                         & \multirow{4}{*}{RGCN}         & \(\text{G}^2\text{A2}\text{C} \)                           & \multirow{4}{*}{0.4959} & \multirow{4}{*}{0.5245} & 0.4907          & 0.5207          & 0.4901          & 0.5200          & 0.4800          & 0.5120           \\
                         &                               & FGA                             &                         &                         & 0.4906          & 0.5203          & 0.4866          & 0.5147          & 0.4700          & 0.5017           \\
                         &                               & RoHe-attack                     &                         &                         & 0.4820          & 0.5143          & 0.4795          & 0.5121          & 0.4539          & 0.4925           \\
                         &                               & HeTa                              &                         &                         & \cellcolor{gray!20}\textbf{0.4055} & \cellcolor{gray!20}\textbf{0.4035} & \cellcolor{gray!20}\textbf{0.382}  &\cellcolor{gray!20} \textbf{0.3726} & \cellcolor{gray!20}\textbf{0.3398} & \cellcolor{gray!20}\textbf{0.3246}  \\ 
\hline
\multirow{16}{*}{DBLP}   & \multirow{4}{*}{HAN}          & \(\text{G}^2\text{A2}\text{C} \)                           & \multirow{4}{*}{0.9239} & \multirow{4}{*}{0.9292} & 0.9210          & 0.9230          & 0.9032          & 0.9089          & 0.8900          & 0.8901           \\
                         &                               & FGA                             &                         &                         & -               & -               & -               & -               & -               & -                \\
                         &                               & RoHe-attack                     &                         &                         & 0.8782          & 0.8838          & 0.8456          & 0.8531          & 0.7636          & 0.7771           \\
                         &                               & HeTa                              &                         &                         & \cellcolor{gray!20}\textbf{0.8236} & \cellcolor{gray!20}\textbf{0.8311} & \cellcolor{gray!20}\textbf{0.5766} & \cellcolor{gray!20}\textbf{0.6021} & \cellcolor{gray!20}\textbf{0.4495} & \cellcolor{gray!20}\textbf{0.4981}  \\ 
\cline{2-11}
                         & \multirow{4}{*}{HGT}          & \(\text{G}^2\text{A2}\text{C} \)                           & \multirow{4}{*}{0.9049} & \multirow{4}{*}{0.9123} & 0.8817          & 0.8876          & 0.8628          & 0.8676          & 0.8066          & 0.8112           \\
                         &                               & FGA                             &                         &                         & -               & -               & -               & -               & -               & -                \\
                         &                               & RoHe-attack                     &                         &                         & 0.8880           & 0.8957          & 0.8608          & 0.8685          & 0.8206          & 0.8281           \\
                         &                               & HeTa                              &                         &                         & \cellcolor{gray!20}\textbf{0.8562} & \cellcolor{gray!20}\textbf{0.8623} & \cellcolor{gray!20}\textbf{0.8076} & \cellcolor{gray!20}\textbf{0.8114} & \cellcolor{gray!20}\textbf{0.7349} & \cellcolor{gray!20}\textbf{0.7357}  \\ 
\cline{2-11}
                         & \multirow{4}{*}{SimpleHGN}    & \(\text{G}^2\text{A}^2\text{C} \)                           & \multirow{4}{*}{0.9232} & \multirow{4}{*}{0.9285} & 0.9219          & 0.9271          & 0.9198          & 0.9250          & 0.9128          & 0.9179           \\
                         &                               & FGA                             &                         &                         & -               & -               & -               & -               & -               & -                \\
                         &                               & RoHe-attack                     &                         &                         & 0.8742          & 0.8799          & 0.8391          & 0.8454          & 0.8010          & 0.8105           \\
                         &                               & HeTa                              &                         &                         & \cellcolor{gray!20}\textbf{0.6687} & \cellcolor{gray!20}\textbf{0.6741} & \cellcolor{gray!20}\textbf{0.6569} & \cellcolor{gray!20}\textbf{0.6604} & \cellcolor{gray!20}\textbf{0.627}  & \cellcolor{gray!20}\textbf{0.6483}  \\ 
\cline{2-11}
                         & \multirow{4}{*}{RGCN}         & \(\text{G}^2\text{A2}\text{C} \)                           & \multirow{4}{*}{0.9023} & \multirow{4}{*}{0.9084} & 0.8891          & 0.8947          & 0.8810          & 0.8862          & 0.8403          & 0.8443           \\
                         &                               & FGA                             &                         &                         & -               & -               & -               & -               & -               & -                \\
                         &                               & RoHe-attack                     &                         &                         & 0.8440          & 0.8510          & 0.7802          & 0.7880          & 0.6007          & 0.6080           \\
                         &                               & HeTa                              &                         &                         & \cellcolor{gray!20}\textbf{0.6611} & \cellcolor{gray!20}\textbf{0.6754} & \cellcolor{gray!20}\textbf{0.5693} & \cellcolor{gray!20}\textbf{0.5933} & \cellcolor{gray!20}\textbf{0.5153} & \cellcolor{gray!20}\textbf{0.5444}  \\ 
\hline
\multirow{16}{*}{ACM}    & \multirow{4}{*}{HAN}          & \(\text{G}^2\text{A2}\text{C} \)                           & \multirow{4}{*}{0.9064} & \multirow{4}{*}{0.9046} & 0.9034          & 0.9043          & 0.9000          & 0.9010          & 0.8912          & 0.8943           \\
                         &                               & FGA                             &                         &                         & 0.8894          & 0.8871          & 0.8706          & 0.8677          & 0.8213          & 0.8168           \\
                         &                               & RoHe-attack                     &                         &                         & 0.8658          & 0.8635          & 0.8253          & 0.822           & 0.7266          & 0.7247           \\
                         &                               & HeTa                              &                         &                         & \cellcolor{gray!20}\textbf{0.4712} & \cellcolor{gray!20}\textbf{0.5576} & \cellcolor{gray!20}\textbf{0.4226} & \cellcolor{gray!20}\textbf{0.5176} & \cellcolor{gray!20}\textbf{0.1818} & \cellcolor{gray!20}\textbf{0.3435}  \\ 
\cline{2-11}
                         & \multirow{4}{*}{HGT}          & \(\text{G}^2\text{A2}\text{C} \)                           & \multirow{4}{*}{0.9087} & \multirow{4}{*}{0.9079} & 0.9078          & 0.9069          & 0.9074          & 0.9065          & 0.9069          & 0.906            \\
                         &                               & FGA                             &                         &                         & 0.8866          & 0.8857          & 0.8590          & 0.8578          & 0.7882          & 0.7865           \\
                         &                               & RoHe-attack                     &                         &                         & 0.8959          & 0.8951          & 0.8825          & 0.8819          & 0.8356          & 0.8352           \\
                         &                               & HeTa                              &                         &                         & \cellcolor{gray!20}\textbf{0.8002} & \cellcolor{gray!20}\textbf{0.8006} & \cellcolor{gray!20}\textbf{0.6941} & \cellcolor{gray!20}\textbf{0.6971} & \cellcolor{gray!20}\textbf{0.682}  & \cellcolor{gray!20}\textbf{0.6883}  \\ 
\cline{2-11}
                         & \multirow{4}{*}{SimpleHGN}    & \(\text{G}^2\text{A2}\text{C} \)                           & \multirow{4}{*}{0.9375} & \multirow{4}{*}{0.9367} & 0.9371          & 0.9362          & 0.9364          & 0.9357          & 0.936           & 0.935            \\
                         &                               & FGA                             &                         &                         & 0.9365          & 0.9357          & 0.9384          & 0.9376          & 0.9296          & 0.9291           \\
                         &                               & RoHe-attack                     &                         &                         & 0.9303          & 0.9296          & 0.9268          & 0.9263          & 0.9154          & 0.915            \\
                         &                               & HeTa                              &                         &                         & \cellcolor{gray!20}\textbf{0.6900} & \cellcolor{gray!20}\textbf{0.7015} & \cellcolor{gray!20}\textbf{0.6689} & \cellcolor{gray!20}\textbf{0.6791} & \cellcolor{gray!20}\textbf{0.1973} &\cellcolor{gray!20} \textbf{0.3510}  \\ 
\cline{2-11}
                         & \multirow{4}{*}{RGCN}         & \(\text{G}^2\text{A2}\text{C} \)                           & \multirow{4}{*}{0.8857} & \multirow{4}{*}{0.8838} & 0.8844          & 0.8824          & 0.8830          & 0.8810          & 0.8807          & 0.8786           \\
                         &                               & FGA                             &                         &                         & 0.8722          & 0.8710          & 0.8654          & 0.8623          & 0.8555          & 0.8512           \\
                         &                               & RoHe-attack                     &                         &                         & 0.8660          & 0.8630          & 0.8410          & 0.8401          & 0.8255          & 0.8205           \\
                         &                               & HeTa                              &                         &                         & \cellcolor{gray!20}\textbf{0.7559} & \cellcolor{gray!20}\textbf{0.7524} &\cellcolor{gray!20} \textbf{0.6605} &\cellcolor{gray!20} \textbf{0.6671} & \cellcolor{gray!20}\textbf{0.5243} & \cellcolor{gray!20}\textbf{0.5551}  \\
\hline
\end{tabular}
}

\caption{Overall attack performance on three datasets across four target HGNN backbones. Lower scores indicate better attacking ability. ``Clean” means the original graphs without perturbations. Vacant positions (“-”) mean that the models run out of memory on large graphs.}
\label{tab:overall_result}
\end{table*}

\begin{figure*}[htbp]
	\centering
	\begin{subfigure}{0.24\linewidth}
		\centering
		\includegraphics[width=0.9\linewidth]{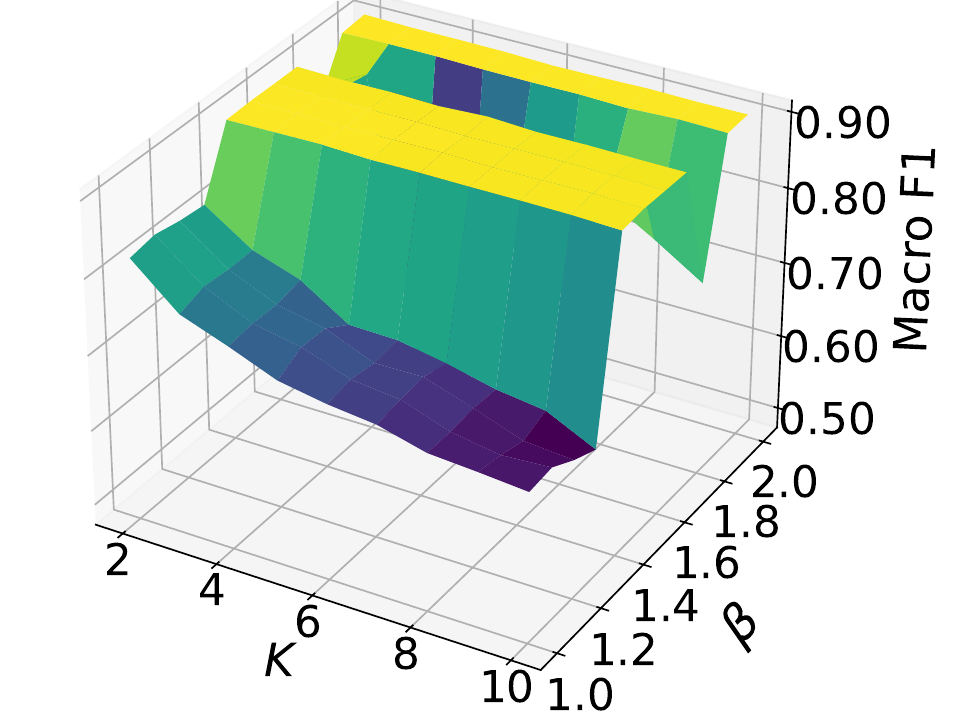}
		\caption{HAN}
		\label{HAN}
	\end{subfigure}
	\centering
	\begin{subfigure}{0.24\linewidth}
		\centering
		\includegraphics[width=0.9\linewidth]{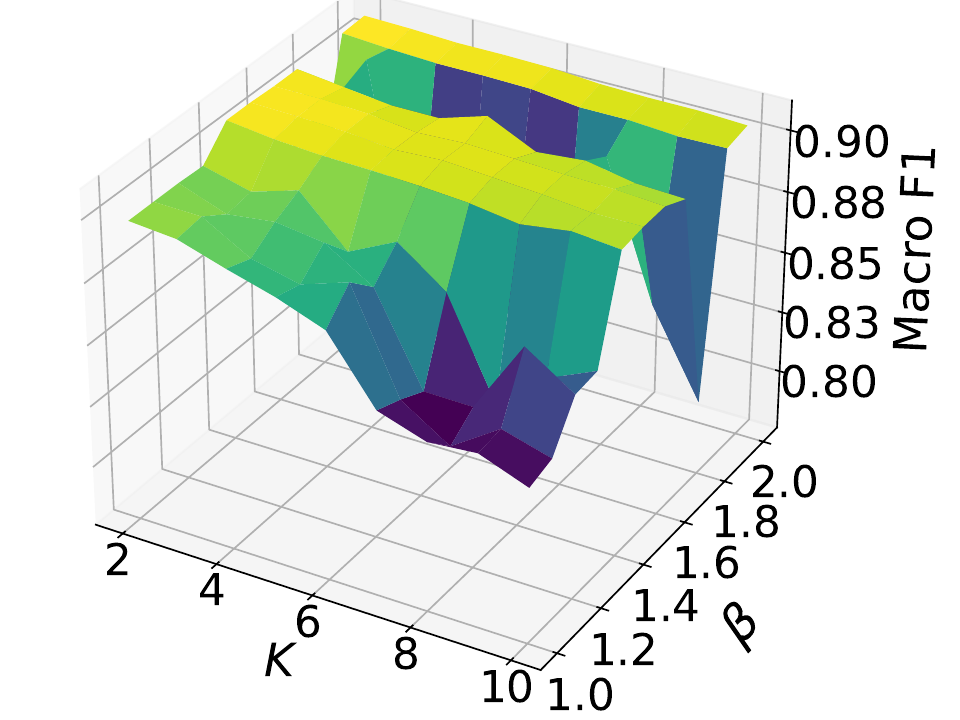}
		\caption{HGT}
		\label{HGT}
	\end{subfigure}
	\centering
	\begin{subfigure}{0.24\linewidth}
		\centering
		\includegraphics[width=0.9\linewidth]{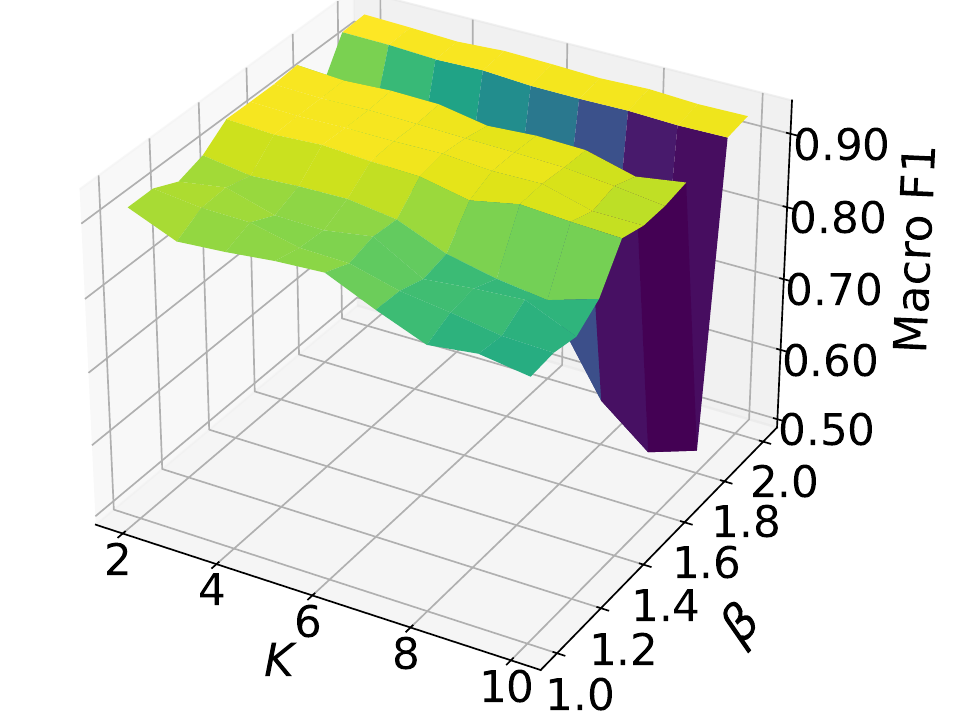}
		\caption{SimpleHGN}
		\label{SimpleHGN}
	\end{subfigure}
        \begin{subfigure}{0.24\linewidth}
		\centering
		\includegraphics[width=0.9\linewidth]{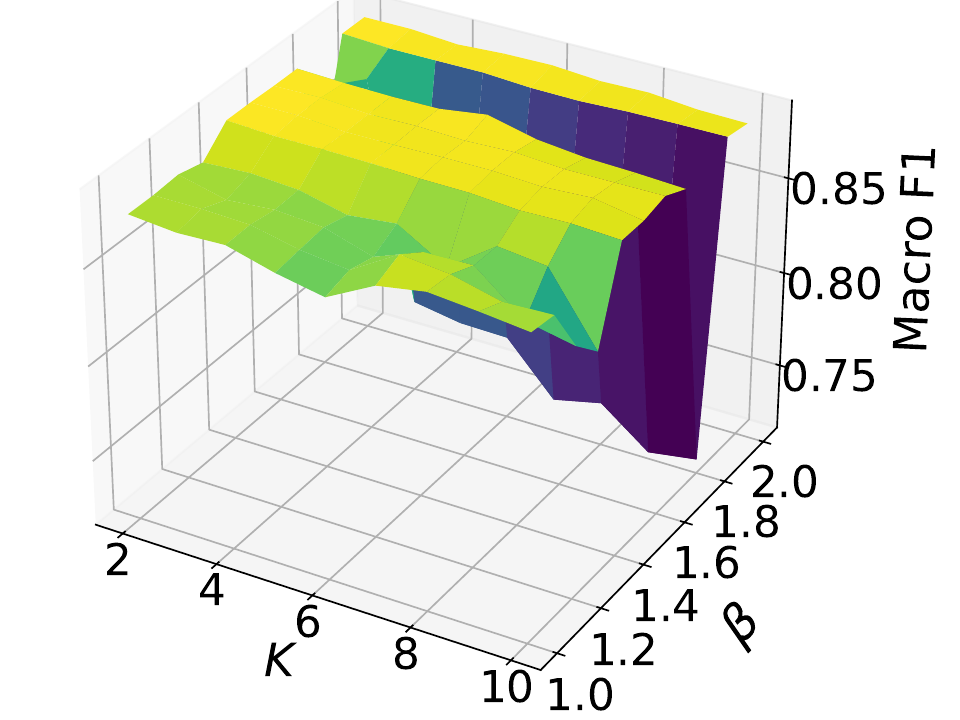}
		\caption{RGCN}
		\label{RGCN}
	\end{subfigure}

	\caption{Analysis of the hyper-parameter \(K\) and \(\beta\) on ACM dataset with an injection rate of 0.01.}
	\label{parameter_both}
\end{figure*}

\subsection{Overall Performance}
We conduct attacks on the surrogate model and use the perturbed graph obtained as the input for the target model, the results are shown in Table \ref{tab:overall_result}. 
We have the following observations: 
\textbf{(1)} \textbf{The proposed HeTa achieves state-of-the-art (SOTA) performance across all datasets and backbones.} This further substantiates the effectiveness of HeTa in HGNN attack. Specifically, with a 0.01 node injection,  compared with the best baseline attack, performance drops are up to 19\% in IMDB, nearly 20\% in DBLP, and almost 35\% in ACM. 
\textbf{(2)} \textbf{Our proposed HeTa can generalize across various target HGNN models.}  Surprisingly, the perturbations trained on the surrogate model achieved SoTa attack across all target models, causing significant performance degradation.
Notably, with a 0.01 node injection, the average performance degradation by nearly 12\% in IMDB, 15\% in DBLP, and 19\% in ACM. 
This generalization is attributed to the unified modeling of different HGNNs based on relational semantics.
\begin{figure}[htbp]
	\centering
	\begin{subfigure}{0.46\linewidth}
		\centering
		\includegraphics[width=0.95\linewidth]{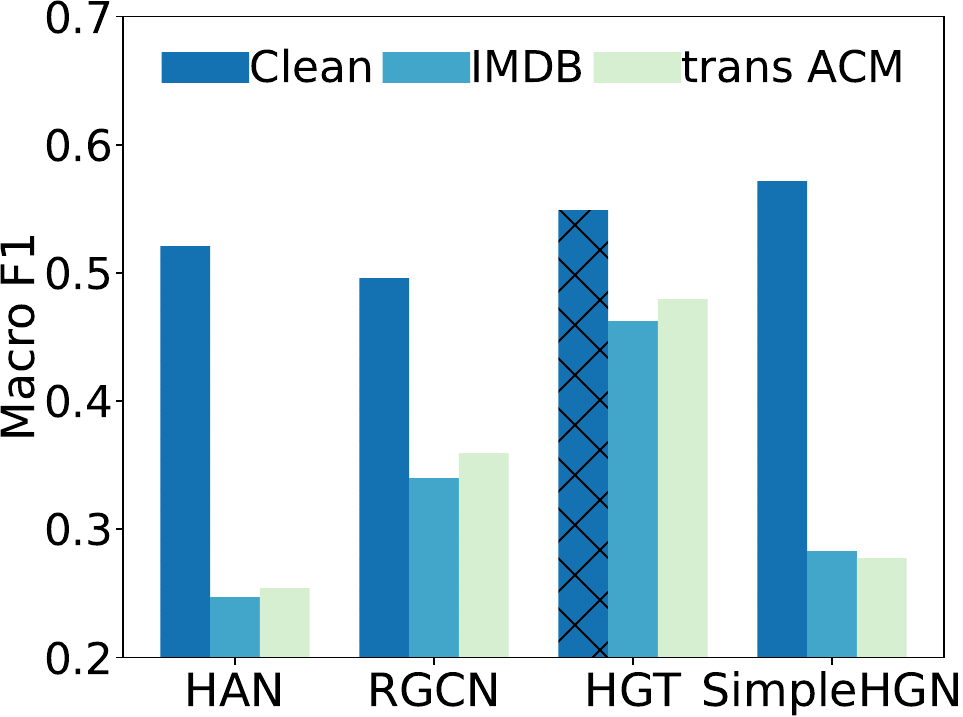}
		\caption{HeTa}
		\label{MY}
	\end{subfigure}
	\centering
	\begin{subfigure}{0.46\linewidth}
		\centering
		\includegraphics[width=0.95\linewidth]{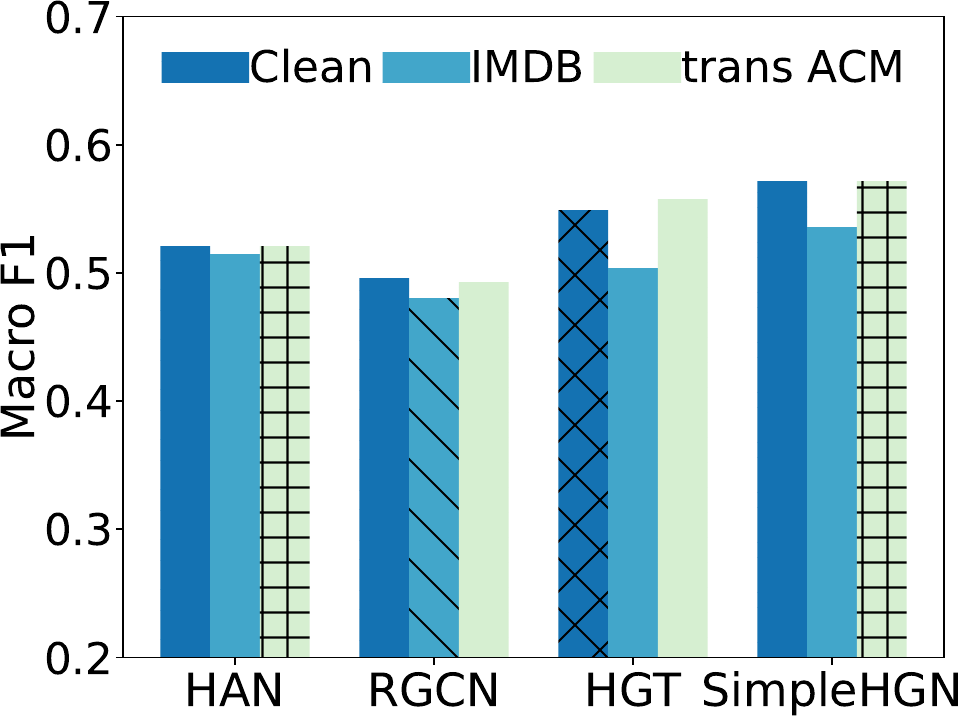}
		\caption{\(\text{G}^2\text{A2C} \)}
		\label{G2A2C}
	\end{subfigure}
	\centering
	\caption{Results on IMDB dataset. `Clean’ indicates without attack, `IMDB’ means attacker trained on IMDB dataset, and `trans ACM’ means attacker trained on ACM  and adapt to IMDB. }
	\label{fig:transfer_data}
\end{figure}
\subsection{Merits of HeTa}
\begin{figure}[htbp]
	\centering
	\begin{subfigure}{0.46\linewidth}
		\centering
		\includegraphics[width=0.95\linewidth]{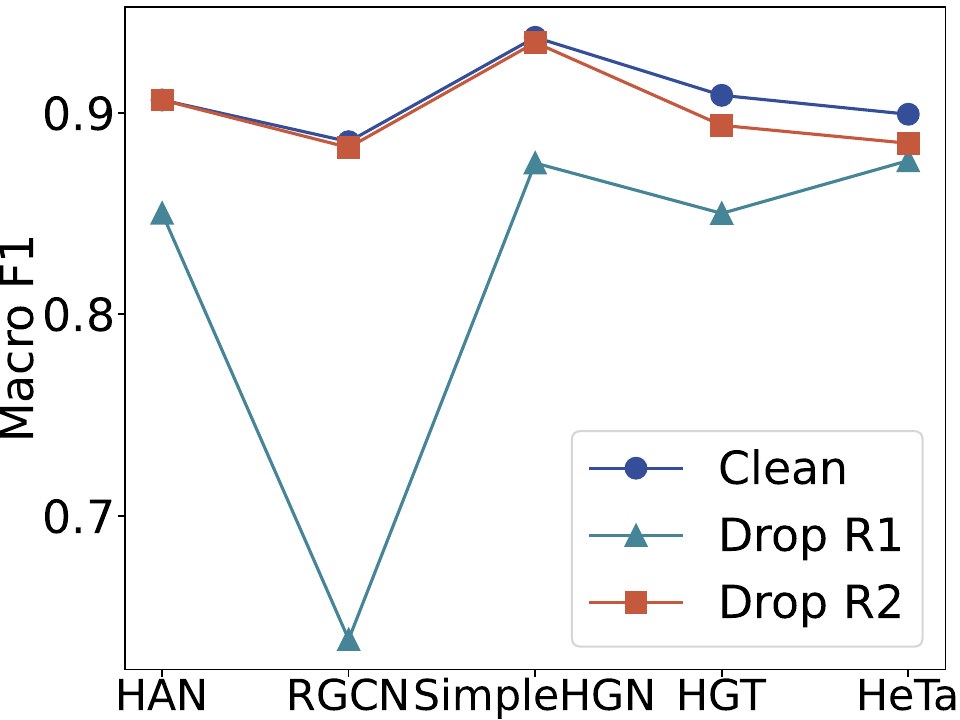}
		\caption{Universality.}
		\label{fig:general}
	\end{subfigure}
	\centering
	\begin{subfigure}{0.46\linewidth}
		\centering
		\includegraphics[width=0.95\linewidth]{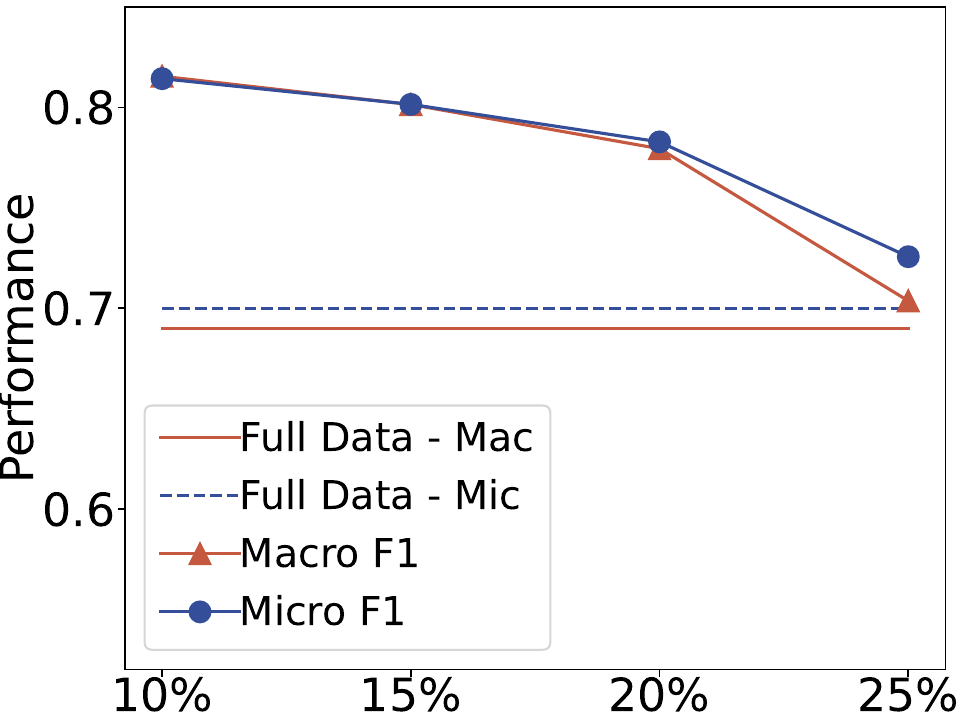}
		\caption{Data Efficiency.}
		\label{de}
	\end{subfigure}
	\centering
	\caption{Results on the ACM dataset at 1\% injection. (a) Dropping relation subgraphs cross five HGNN models. (b)  Sampling training sets at different ratios.}
	\label{fig:total}
\end{figure}

\textbf{HeTa achieves cross-graph transferability.}
We pretrain the fake node generator in HeTa on the ACM dataset, freeze its parameters, and only fine-tune the surrogate model to adapt to the IMDB dataset. For \(\text{G}^2\text{A2}\text{C} \), we also freeze its generator and finetuning with its own loss. The results in Figure \ref{fig:transfer_data} (and additional results in Figure \ref{fig:transfer_data} in the Appendix) show that the performance of trans ACM is close to that of IMDB, particularly on RGCN and HGT, where they are nearly identical. 
HeTa shows strong cross-graph transferability, this is because our attacker has learned universal characteristics within HGs, enabling it to efficiently adapt to new graph distributions.\\
\textbf{Universality of our foundation surrogate model.} 
We conduct experiments to validate the effect of dropping relation subgraphs with varying importance assigned by $\boldsymbol{\mu}$ in Figure \ref{fig:general}.
According to  $\boldsymbol{\mu}$, it indicates that R1 (author-paper) \(>\) R2 (paper-term). 
The results show that removing R2 has little impact, while removing R1 causes a significant performance drop across all HGNNs. Notably, RGCN's performance drops sharply after removing R1, as it relies heavily on R1. This suggests that HeTa can identify the unified importance distribution of relations across various HGNNs.\\
\textbf{HeTa exhibits strong data efficiency.} 
We sample a small subset of nodes from the full training data at different ratios to create a new dataset, as shown in Figure \ref{de}.
Thanks to the minimal number of parameters in our surrogate model, which enables efficient training use of limited data. Specifically, when using only 25\% of the training data, the attacker already achieves performance comparable to that with the full dataset, highlighting its data efficiency.\\
\begin{figure}[htbp]
	\centering
	\begin{subfigure}{0.46\linewidth}
		\centering
		\includegraphics[width=0.95\linewidth]{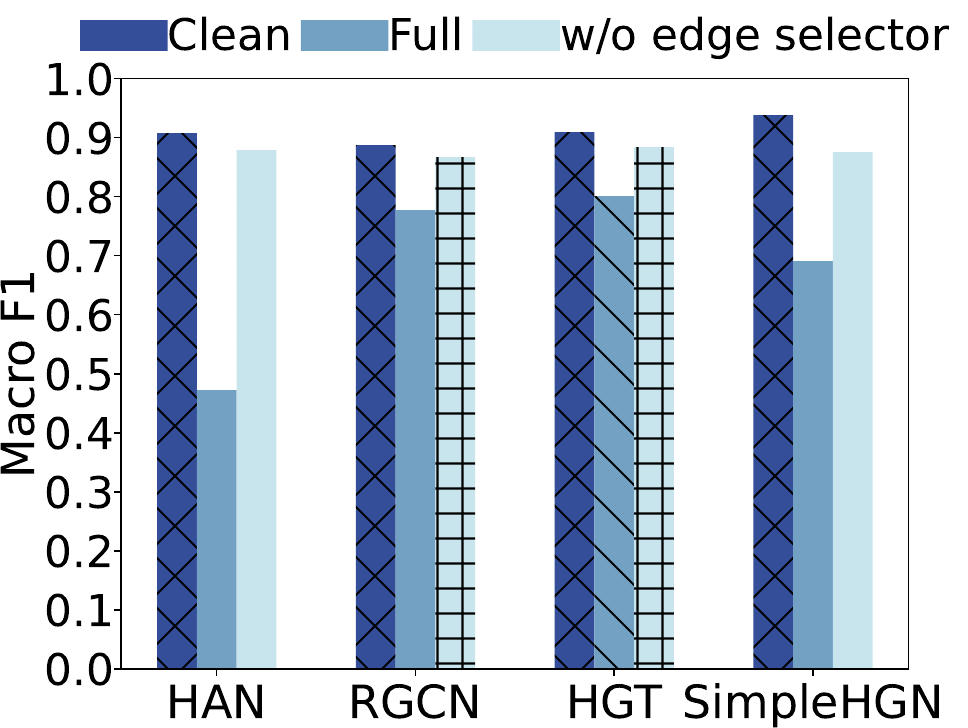}
		\caption{ACM}
		\label{HAN}
	\end{subfigure}
	\centering
	\begin{subfigure}{0.46\linewidth}
		\centering
		\includegraphics[width=0.95\linewidth]{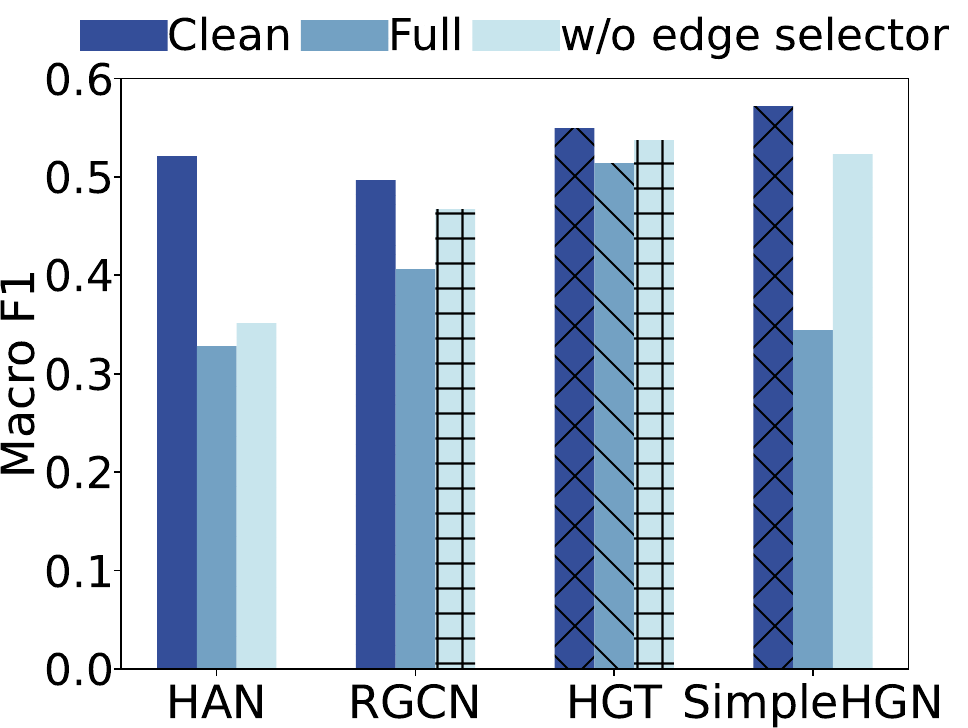}
		\caption{IMDB}
		\label{HGT}
	\end{subfigure}
	\centering
	\caption{Ablation study with an injection rate of 0.01.}
	\label{ablation}
\end{figure}
\subsection{Model Analysis}
\textbf{Ablation Study.} 
Since a key design of HeTa is the selection of fake edges, we replace this strategy with random edge selection at an injection rate of 0.01, as shown in Figure \ref{ablation}.
From the result, the full method achieves the best performance, demonstrating that our key design can significantly degrade the performance of the target model.\\
\textbf{Hyper-parameter Study.} We analyze the impact of the injection degree \(K\) and the penalty term \(\beta\) in Figure \ref{parameter_both}. \(K\) controls the influence range of injected nodes, and \(\beta\) guides the selection of attack relations. 
Smaller \(K\)  enhance attack effects by carefully selecting neighbors within a limited influence range, maintaining low visibility. However, larger \(K\) values can enhance attacks by broadening the influence. Attack effects initially improve with \(\beta\), peaking at \(\beta=1.8\), and then deteriorate due to excessive punishment causing attack dispersion. More results are presented in Appendix D.1.
\subsection{Conclusion}
In this study, we introduce the foundation attack model within HGNNs, HeTa, which identifies shared attack patterns based on relational semantics, thereby enabling the attack process on HGs to generalize.
We design a lightweight surrogate model to simplify various HGNN propagation mechanisms and model the important distribution of shared relational semantic units. 
Then we propose a relation-by-relation serialized attack process, which involves injecting adversarial nodes into each relation and generating fake edges.
Experiments conducted on three datasets across four HGNN backbones have demonstrated that our method not only outperforms existing attack models but also exhibits remarkable generalization ability.

\newpage

\section*{Acknowledgements}
This work was supported in part by the Zhejiang Provincial Natural Science Foundation of China under Grant LDT23F01015F01, in part by the Key Technology Research and Development Program of the Zhejiang Province under Grant No. 2025C1023 and in part by the National Natural Science Foundation of China (No. 62372146, 62322203, 62172052).



\bibliographystyle{named}
\bibliography{ijcai25}

\clearpage

\appendix
\clearpage


\begin{algorithm}[tb]
    \caption{The Attack Process in HeTa.}
    \label{alg:algorithm}
    \textbf{Input}: Clean graph $\mathcal{G}$,
    the attack units $\mathcal{A} = \{A_r,r \in \mathcal{R}\}$ and relation weight $\boldsymbol{\mu}=\{\mu_{1},\mu_{2}...\mu_{|\mathcal{R}|}\}$ from surrogate model. 
    \\
    \textbf{Parameter}: Budget on injected rate \(\rho\) and the degree of each injected node $K$. The threshold for the total number of injected nodes  $M$.\\
    \textbf{Output}:Attacked graph $\mathcal{G'}=\{\mathcal{A'},\mathcal{F'}\}$.
    
    \begin{algorithmic}[1]
    \FOR { m =1 to M}
        \STATE Select the current relation unit to attack based on Eq~\eqref{eq:relation_select}.
         \STATE Initialize the features and connection status of the injected nodes $v_{in}$.
            \STATE Generate fake node feature $x_{in}^m$ according Eq.\eqref{node_generator}.
            \STATE Select fake edge according Eq.\eqref{fake_edge_selector}.
            \STATE Update the relational weight coefficients $ \mu_{r}^{m+1} = \frac{\mu_{r}^{m}}{\beta}$.
    \ENDFOR
    \RETURN $\mathcal{G'}=\{\mathcal{A'},\mathcal{F'}\}$
\end{algorithmic}
\end{algorithm}
\begin{table}
\small
\setlength{\tabcolsep}{3pt}
\centering
\begin{tabular}{ccccccc} 
\hline
\begin{tabular}[c]{@{}c@{}}Dataset\end{tabular} & \#Nodes & \begin{tabular}[c]{@{}c@{}}\#Node\\~Types\end{tabular} & \#Edges & \begin{tabular}[c]{@{}c@{}}\#Edge\\Types\end{tabular} & Target & \#Classes  \\ 
\hline
DBLP                                                         & 26128   & 4                                                      & 239566  & 6                                                     & author & 4          \\
ACM                                                          & 10942   & 4                                                      & 547872  & 8                                                     & paper  & 3          \\
IMDB                                                         & 12772   & 3                                                      & 37288   & 4                                                     & movie  & 4           \\
\hline
\end{tabular}
\caption{Statistics of datasets.}
\label{tab:dataset}
\end{table}
\section{Additional Analysis}\label{app:Additional_Analysis}
Since we choose the relation with the largest weight in the surrogate model to attack, we focus only on the attacking relation \(r\) in this analysis and we compute the gradient in forward propagation \cite{wang2020scalable}.
\begin{align}
    A'_r = \begin{bmatrix}
            A_r & e_r \\
            e_r^\top & 0
        \end{bmatrix},
    D'_r = \begin{bmatrix}
            D_r+d_1 & 0 \\
             0 & d_2
        \end{bmatrix},
\end{align}%
where \(A'_r \in \mathbb{R}^{|N+1| \times |N+1|} \) denotes the adjacency matrix obtained after attack and \(D'_r\in \mathbb{R}^{|N+1| \times |N+1|} \) denotes the degree matrix of \(A_r^{'}\). \(A_r\) denotes the adjacency matrix of a relation in the original graph, \(D_r\) denotes the degree matrix of the original graph. Since the attacks are performed sequentially, the result of the previous injection will have an impact on the subsequent injections, so the degree change of the original graph needs to be taken into account. Thus \(d_1\in \mathbb{R}^{N \times 1} \) is the degree of the fake node with existing node and \(d_2\) is the degree of injected node (set as \(K\) here). The vector \(e_r\in \mathbb{R}^{N \times 1} \)  represents the connection between the injected node and the original graph, with the elements in \{0, 1\}, where 0 means no connecting edge and 1 otherwise.  
\begin{align}
   D'_r = \begin{bmatrix} D_r + d_1 + I & 0 \\ 0 & d_2 + 1 \end{bmatrix},
   \tilde{A'}_r = \begin{bmatrix}
    \tilde{A}_r & e_r \\
        e_r^\top & I
        \end{bmatrix}.
\end{align}%

Thus, let \(\lambda=D_r + d_1 + I\), \(d_r=d_2+1\), the symmetric normalization of the adjacency matrix can be expressed as:
\begin{align}
    \hat{A'}_r = \tilde{D}_r^{-\frac{1}{2}} \tilde{A'}_r \tilde{D}_r^{-\frac{1}{2}} = \begin{bmatrix}
\lambda^{-\frac{1}{2}} \tilde{A}_r \lambda^{-\frac{1}{2}} & \lambda^{-\frac{1}{2}} e_r d_r^{-\frac{1}{2}} \\
d_r^{-\frac{1}{2}} e_r^T \lambda^{-\frac{1}{2}} & d_r^{-1}
\end{bmatrix}.
\end{align}%
To clarify, we define \(GD=\lambda^{-\frac{1}{2}} \tilde{A}_r \lambda^{-\frac{1}{2}}\) and \(\hat{e}_r=\lambda^{-\frac{1}{2}} e_r d_r^{-\frac{1}{2}}\):
\begin{align}
    \hat{A'}_r = \begin{bmatrix}
    GD & \hat{e}_r\\
    \hat{e}_r^T & d_r^{-1}\\
\end{bmatrix},
\end{align}%
\begin{align}
\hat{A_r'}^2 
= \begin{bmatrix} GD^2 + \hat{e}_r \hat{e}_r^T & GD \hat{e}_r + \hat{e}_r d_r^{-1} \\ \hat{e}_r^T GD + d_r^{-1} \hat{e}_r^T & \hat{e}_r^T \hat{e}_r + d_r^{-2} \end{bmatrix}.
\end{align}%

We first linearize the two-layer surrogate model formula in Eq.\eqref{eq:Z_softmax}, \(Z_r' = \hat{A'_r}^2 F' W\), 
 with features $F'$ after injected. 
In our work, the learnable parameters in surrogate model in Eq.\eqref{eq:projector} denote as $W_1$ and the learnable parameters in Eq.\eqref{eq:Z_softmax} denote as \(W_2\). Thus two-layer surrogate model can be linearize as \(Z_r' = \hat{A'_r}^2 F'W_1 W_2=\hat{A'_r}^2 H' W_2\), given by  
$H'=\begin{bmatrix}H\\h_{in} \end{bmatrix}$ where $H$ and $h_{in}$ are alignment feature as in Eq.\eqref{eq:projector}. 
We linearize $\mathcal{L}_{cw}$:
\begin{align}
      \mathcal{L}_{cw} = [\hat{A'}^2H'W_2]_{v,c}-\max_{y_v \neq c}[\hat{A'}^2H'W_2]_{v,y_v},
\end{align}%
where \([\cdot]_{i,j}\) is \((i,j)\) entry of the matrix. Since \(Z_r'\) is linear with injected edge \(e_r\) and \(\mathcal{L}_{cw}\) is also linear with \(Z_r'\), we can obtain \(e_r\)'s gradient:
\begin{align}
    \frac{\partial \mathcal{L}_{cw}}{\partial e_{{r}}}
    =& \frac{\partial}{\partial e_r} \left[ \left( GD^2 + \hat{e}_r \hat{e}_r^T \right) H  + \left( GD \hat{e}_r + \hat{e}_r d_r^{-1} \right) h_{in}  \right] W_2\nonumber\\
    =& d_r^{-\frac{1}{2}} \lambda^{-\frac{1}{2}} \bigg[ d_r^{-\frac{1}{2}} H W_2 \ Diag(\lambda^{-\frac{1}{2}} e_r) \nonumber \\
    & \quad + [GD]_{:,\mathcal{I}_{V}} \ Diag(h_{in} W_2) \bigg],
\end{align}
where $\mathcal{I}_{V}$ is index set for test nodes. In order to prevent a gradientless situation when the initial fake nodes are not connected to the original graph, we initialize the fake nodes to be connected to all \(\phi_{con}(r)\) nodes. Specifically, we set $[e_r]_{\mathcal{I}(\phi_{con}(r))} = 1$, where $\mathcal{I}(\phi_{con}(r))$ denotes the index of the $\phi_{con}(r)$ node in $\mathcal{G} $ in this work. Therefore:
\begin{align}
    \frac{\partial \mathcal{L}_{cw}}{\partial e_{{r}}}=&HW_2 + \frac{[GD]_{:,\mathcal{I}_{V}} \ Diag(h_{in}W_2)}{\sqrt{d_r} \sqrt{\lambda}}.
\end{align}
The first term can be ignored since it does not contain fake node. We then analyze and simplify the second term as: 
\begin{equation}
\lambda^{-\frac{3}{2}}d_r^{-\frac{1}{2}} (A'_r+I),
\end{equation}
where the degree of injected node \(d_r\) is set as \(K+1\) here. we can observe that the smaller \(\lambda\) is, the gradient is larger and the more a node may be attacked.

 \begin{figure*}[htbp]
	\centering
	\begin{subfigure}{0.24\linewidth}
		\centering
		\includegraphics[width=0.9\linewidth]{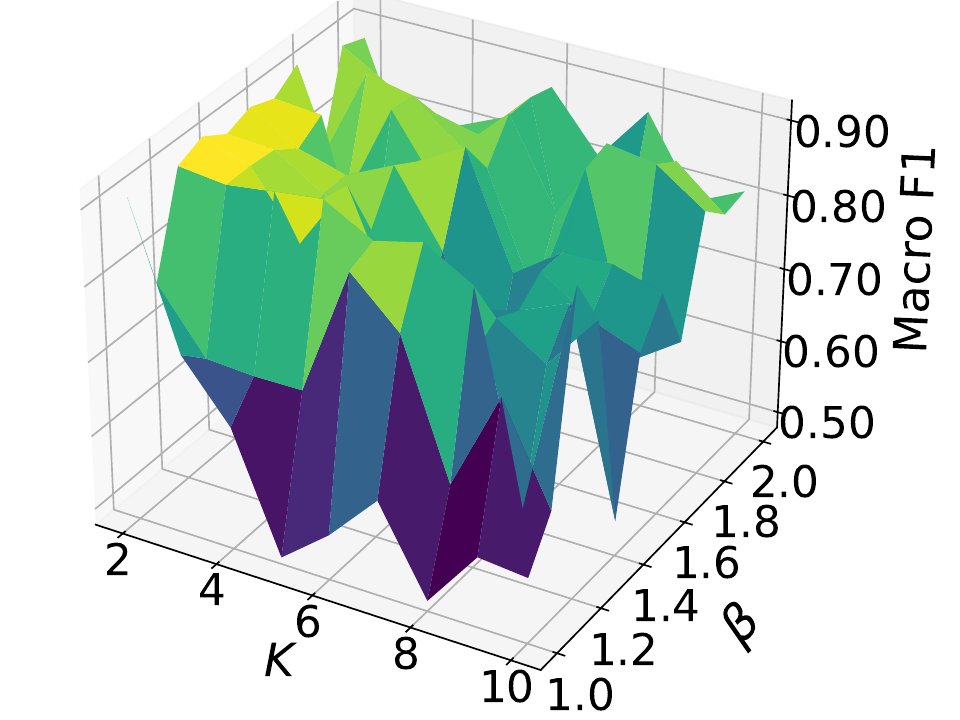}
		\caption{HAN}
		\label{HAN}
	\end{subfigure}
	\centering
	\begin{subfigure}{0.24\linewidth}
		\centering
		\includegraphics[width=0.9\linewidth]{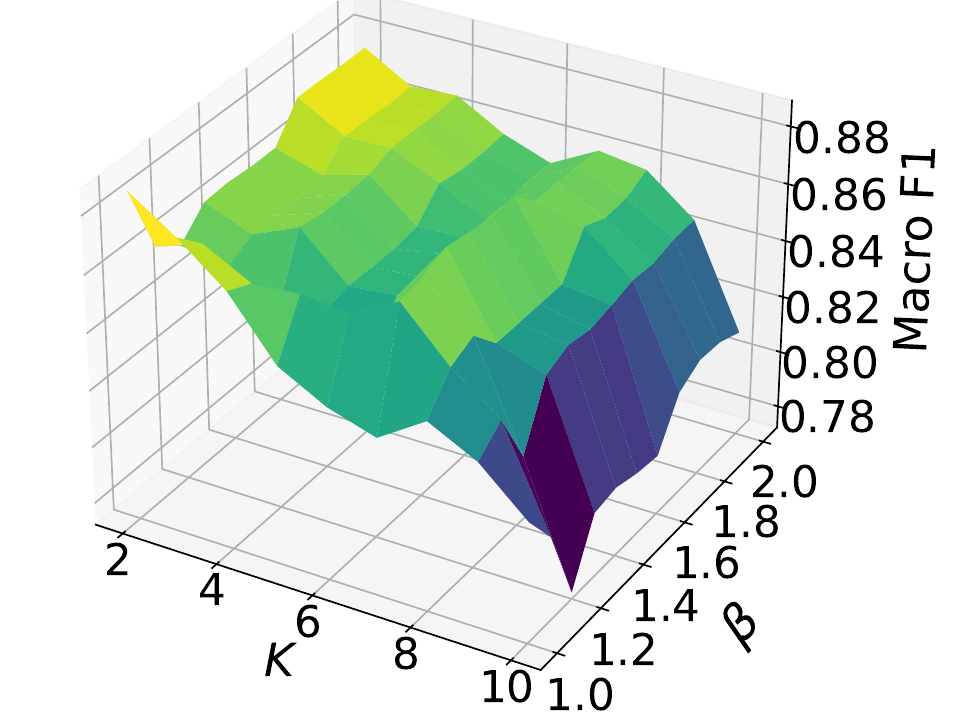}
		\caption{HGT}
		\label{HGT}
	\end{subfigure}
	\centering
	\begin{subfigure}{0.24\linewidth}
		\centering
		\includegraphics[width=0.9\linewidth]{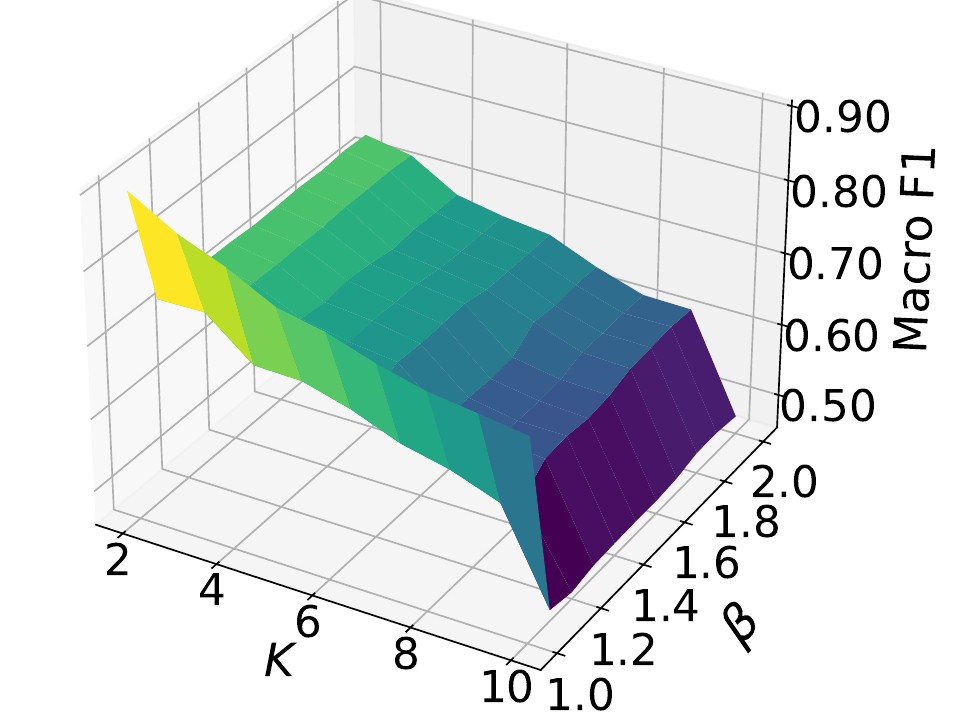}
		\caption{SimpleHGN}
		\label{SimpleHGN}
	\end{subfigure}
        \begin{subfigure}{0.24\linewidth}
		\centering
		\includegraphics[width=0.9\linewidth]{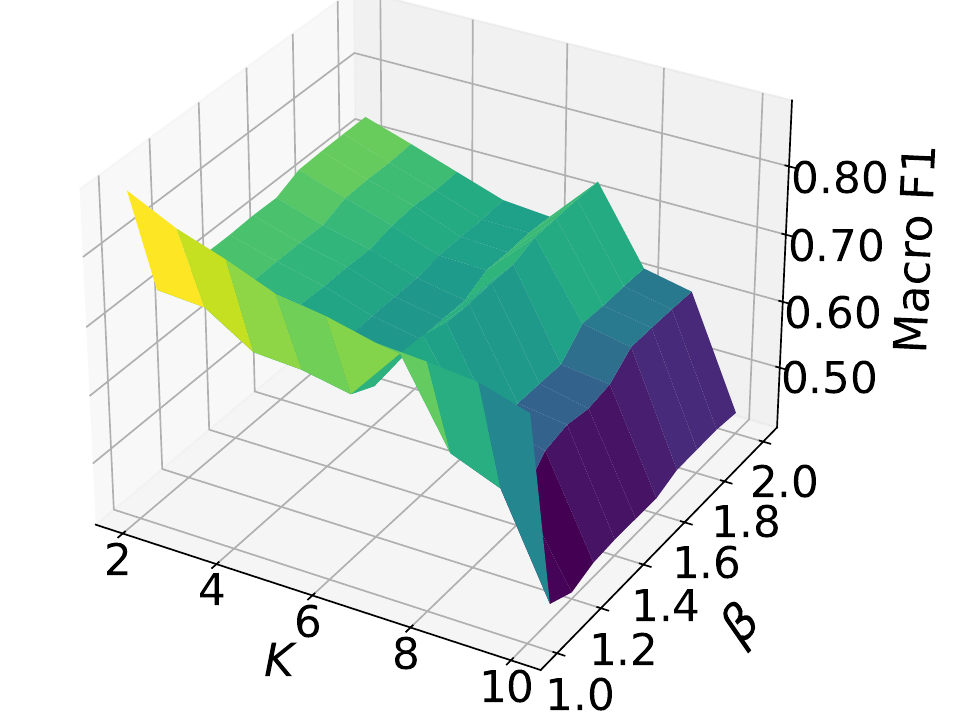}
		\caption{RGCN}
		\label{RGCN}
	\end{subfigure}
\begin{subfigure}{0.24\linewidth}
		\centering
		\includegraphics[width=0.9\linewidth]{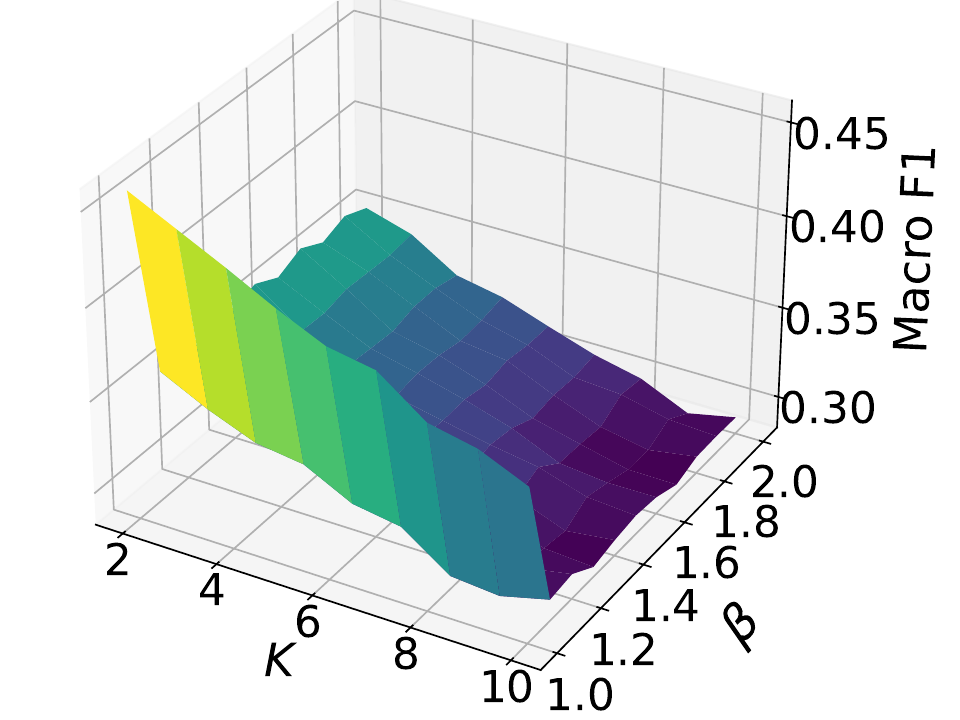}
		\caption{HAN}
		\label{HAN}
	\end{subfigure}
	\centering
	\begin{subfigure}{0.24\linewidth}
		\centering
		\includegraphics[width=0.9\linewidth]{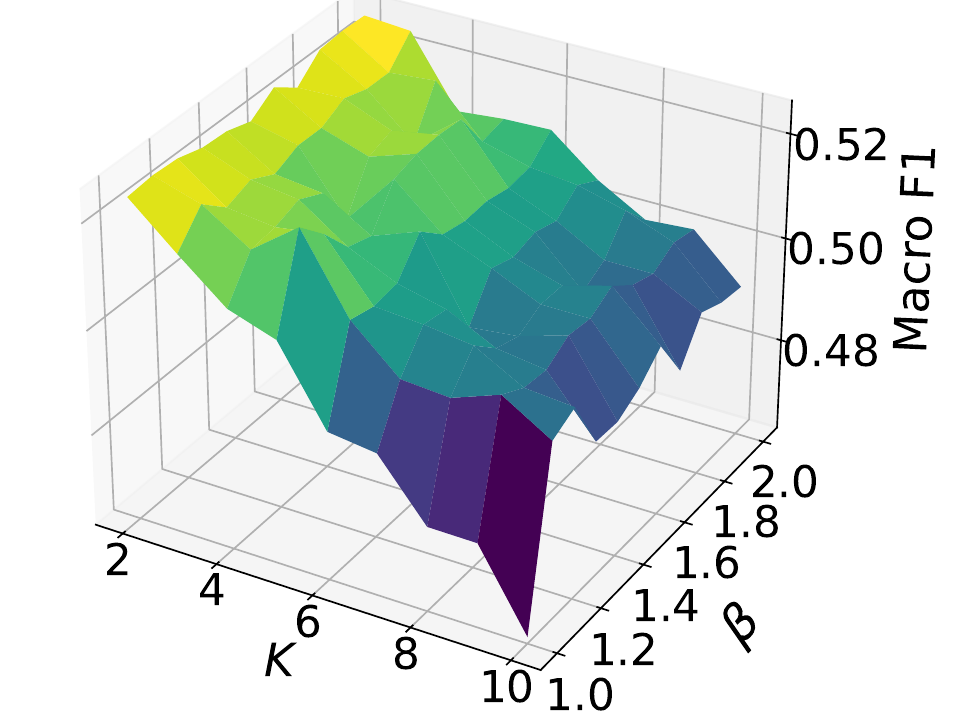}
		\caption{HGT}
		\label{HGT}
	\end{subfigure}
	\centering
	\begin{subfigure}{0.24\linewidth}
		\centering
		\includegraphics[width=0.9\linewidth]{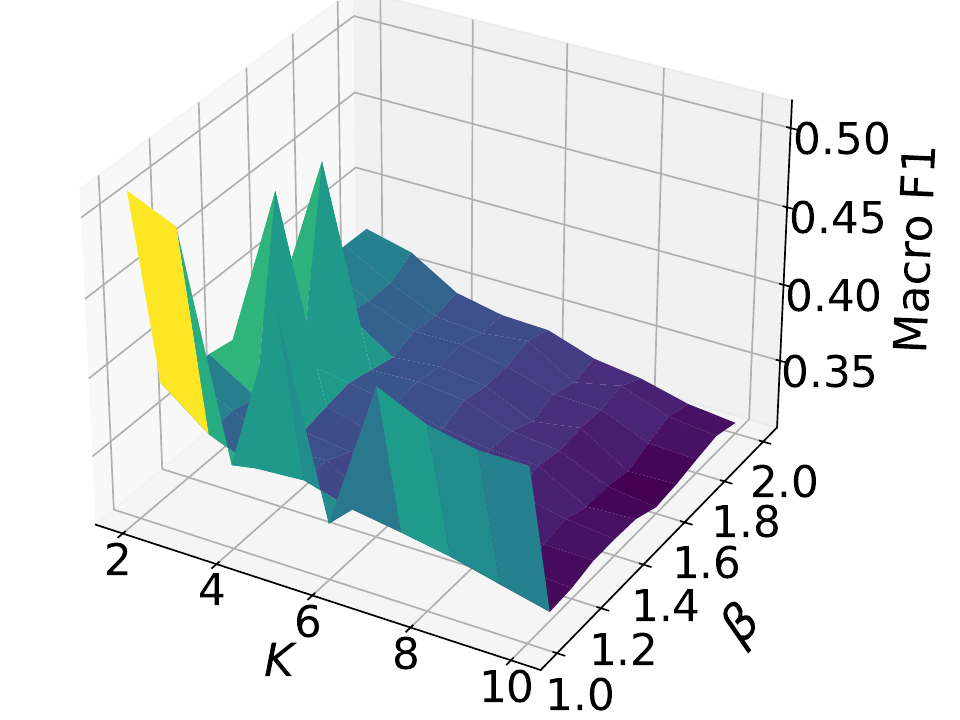}
		\caption{SimpleHGN}
		\label{SimpleHGN}
	\end{subfigure}
        \begin{subfigure}{0.24\linewidth}
		\centering
		\includegraphics[width=0.9\linewidth]{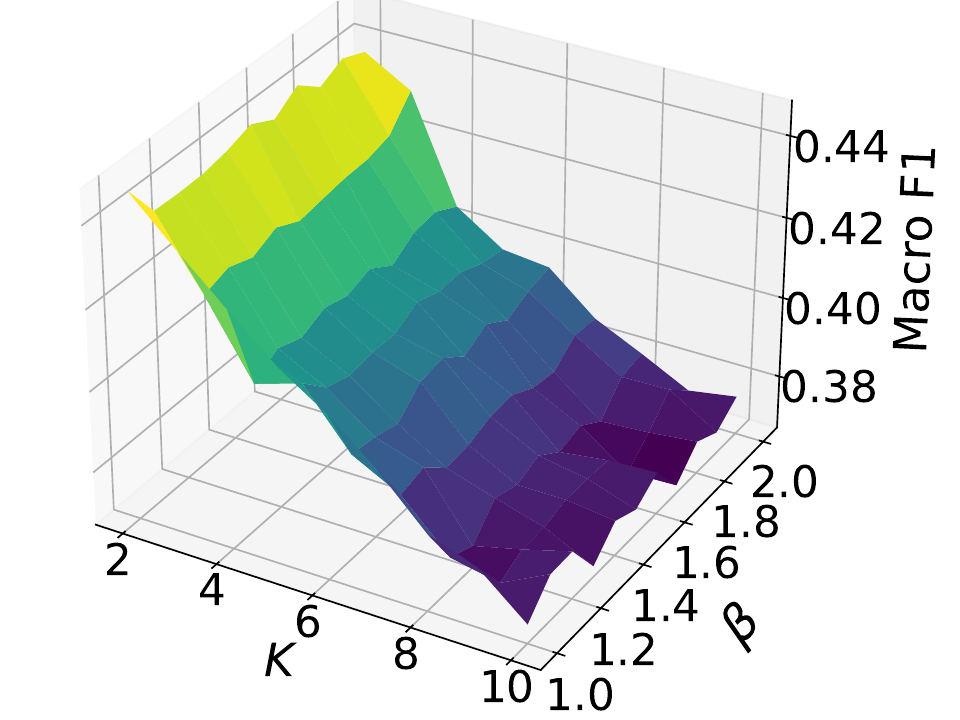}
		\caption{RGCN}
		\label{RGCN}
	\end{subfigure}
	\caption{Analysis of the hyper-parameter \(K\) and \(\beta\) on DBLP, IMDB dataset with an injection rate of 0.01.}
	\label{parameter_both_appendix}
\end{figure*}

\begin{table*}[h!]
\small
\centering
\begin{tabular}{c|c|cc|cc|cc} 
\hline
\multicolumn{2}{c|}{Dataset}               & \multicolumn{2}{c|}{IMDB}         & \multicolumn{2}{c|}{DBLP}         & \multicolumn{2}{c}{ACM}           \\ 
\hline
Target Model               & Attack Methods & Macro F1        & Micro F1        & Macro F1        & Micro F1        & Macro F1        & Micro F1         \\ 
\hline
\multirow{3}{*}{HAN}       
                           & RoHe-attack    & 0.4583          & 0.4955          & \uline{0.7636}  & \uline{0.7771}  & 0.7266          & 0.7247           \\
                           & HeTa         & \uline{0.2465}  & \textbf{0.2879} & \textbf{0.4495} & \textbf{0.4981} & \textbf{0.1818} & \textbf{0.3435}  \\
                           & Simple         & \textbf{0.2437} & \uline{0.3069}  & 0.9074          & 0.9126          & \uline{0.3057}  & \uline{0.4164}   \\ 
\hline
\multirow{3}{*}{HGT}       
                           & RoHe-attack    & 0.5053          & 0.528           & 0.8206          & 0.8281          & 0.8356          & 0.8352           \\
                           & HeTa         & \textbf{0.4624} & \textbf{0.4601} & \uline{0.7349}  & \uline{0.7357}  & \uline{0.682}   & \uline{0.6883}   \\
                           & Simple         & \uline{0.5009}  & \uline{0.5132}  & \textbf{0.6802} & \textbf{0.7158} & \textbf{0.6078} & \textbf{0.6165}  \\ 
\hline
\multirow{3}{*}{SimpleHGN} 
                           & RoHe-attack    & 0.3951          & 0.4023          & 0.801           & 0.8105          & 0.9154          & 0.915            \\
                           & HeTa         & \textbf{0.2825} & \textbf{0.3511} & \uline{0.627}   & \uline{0.6483}  & \textbf{0.1973} & \textbf{0.351}   \\
                           & Simple         & \uline{0.3948}  & \uline{0.3745}  & \textbf{0.2634} & \textbf{0.3693} & \uline{0.5314}  & \uline{0.5632}   \\ 
\hline
\multirow{3}{*}{RGCN}      
                           & RoHe-attack    & 0.4539          & 0.4925          & 0.6007          & 0.608           & 0.8255          & 0.8205           \\
                           & HeTa         & \textbf{0.3398} & \textbf{0.3246} & \uline{0.5153}  & \uline{0.5444}  & \uline{0.5243}  & \uline{0.5551}   \\
                           & Simple         & \uline{0.4072}  & \uline{0.4072}  & \textbf{0.2529} & \textbf{0.3651} & \textbf{0.4432} & \textbf{0.4976}  \\
\hline
\end{tabular}
\caption{Attack results for the three datasets with node injection 0.05. Lower scores indicate better attacking ability. `Simple’ means targeting nodes with lower degrees. The best results are highlighted in bold. The second best results are indicated with underlines.}
\label{tab:Simple_Model_Efficiency}
\end{table*}

\section{Details of Experimental Settings} 
\subsection{Dataset}\label{app:dataset}
The node features of ACM and DBLP are represented using a bag-of-words approach, and one-hot vectors are assigned to nodes that lack initial features. The statistics for these datasets are presented in Table  \ref{tab:dataset}.

\subsection{Details of baselines}\label{app:attack_method_detail}
\begin{itemize}
\item 
FGA~\cite{chen2018fast}: is a targeted evasion attack that uses the gradient of the model's loss  with respect to the input to determine the direction of the perturbation, thereby attack the structure of the original graph.
\item 
RoHe-attack~\cite{zhang2022robust}: performs attacks on specific relations by transforming the HG into a homogeneous graph using meta-path extraction. 

\item \(\text{G}^2\text{A2}\text{C} \)~\cite{ju2023let}:  is a targeted evasion attack that uses reinforcement learning to generate malicious nodes.
\end{itemize}

The publicly available implementations of the models used in this study can be accessed at the following URLs:
\begin{itemize}
    \item \textbf{HGNN} 
    \begin{itemize}
        \item HAN, HGT, SimpleHGN, RGCN: \url{https://github.com/BUPT-GAMMA/OpenHGNN}
    \end{itemize}
    \item \textbf{Attack model}
    \begin{itemize}
        \item FGA: \url{https://github.com/DSE-MSU/DeepRobust}
    \end{itemize}
    \begin{itemize}
        \item \(\text{G}^2\text{A2}\text{C} \): \url{https://github.com/jumxglhf/G2A2C}
    \end{itemize}
    
\end{itemize}
\begin{figure}[htbp]
	\centering
	\begin{subfigure}{0.46\linewidth}
		\centering
		\includegraphics[width=0.95\linewidth]{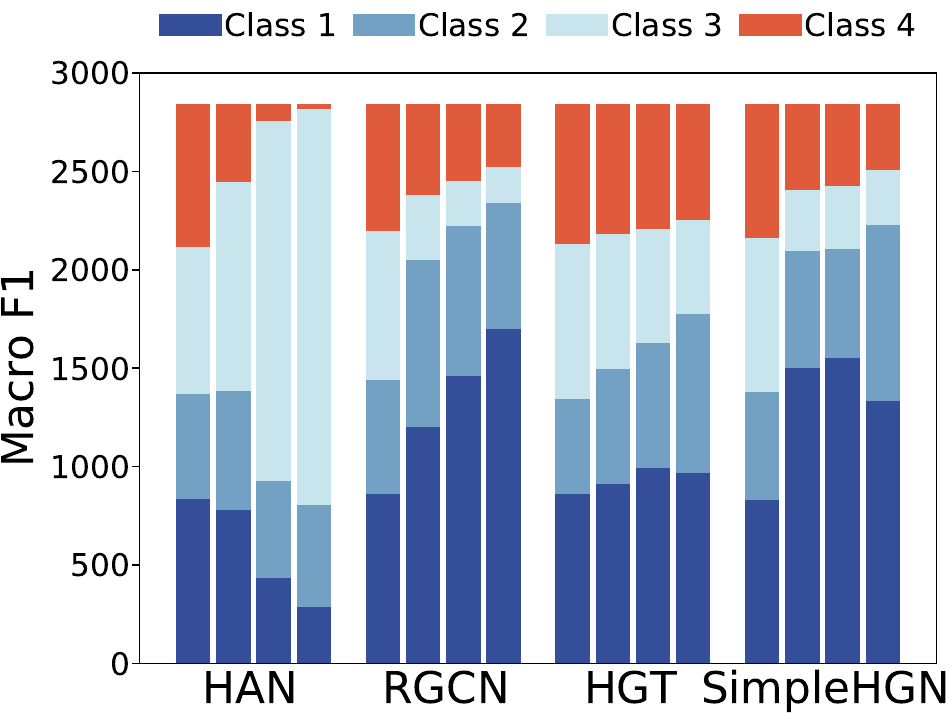}
		\caption{DBLP}
		\label{HAN}
	\end{subfigure}
	\centering
	\begin{subfigure}{0.46\linewidth}
		\centering
		\includegraphics[width=0.95\linewidth]{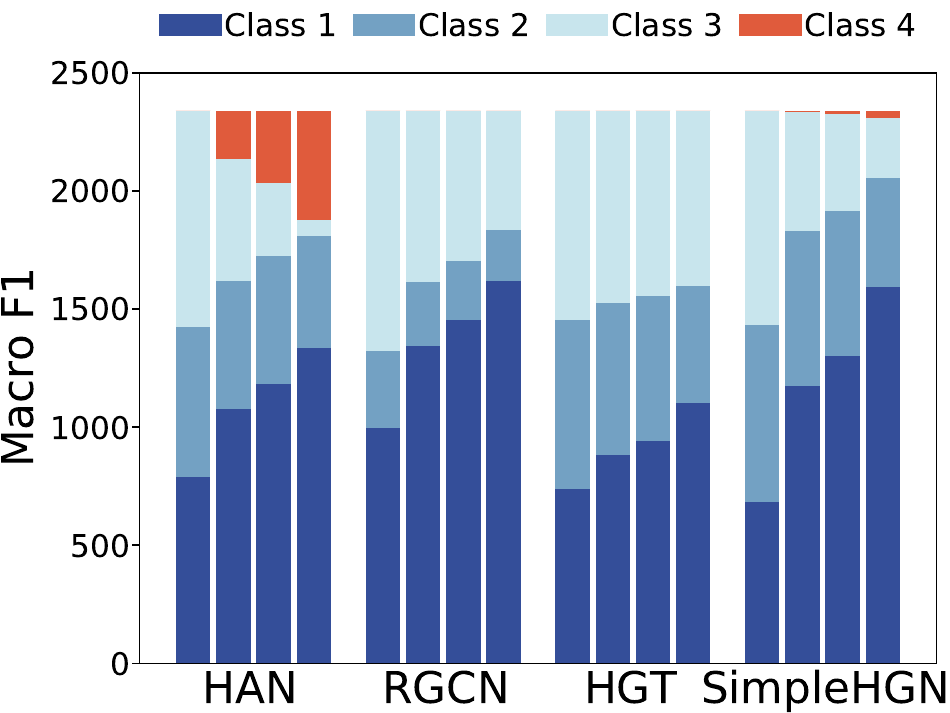}
		\caption{IMDB}
		\label{HGT}
	\end{subfigure}
        \centering
	\centering
	\caption{Label distribution across four target models under varying node injection rates (clean, 1\%, 2\%, 5\%) on DBLP, IMDB.}
	\label{fig:label_distribution}
\end{figure}
\begin{figure*}[h]
	\centering
	\begin{subfigure}{0.24\linewidth}
		\centering
		\includegraphics[width=0.9\linewidth]{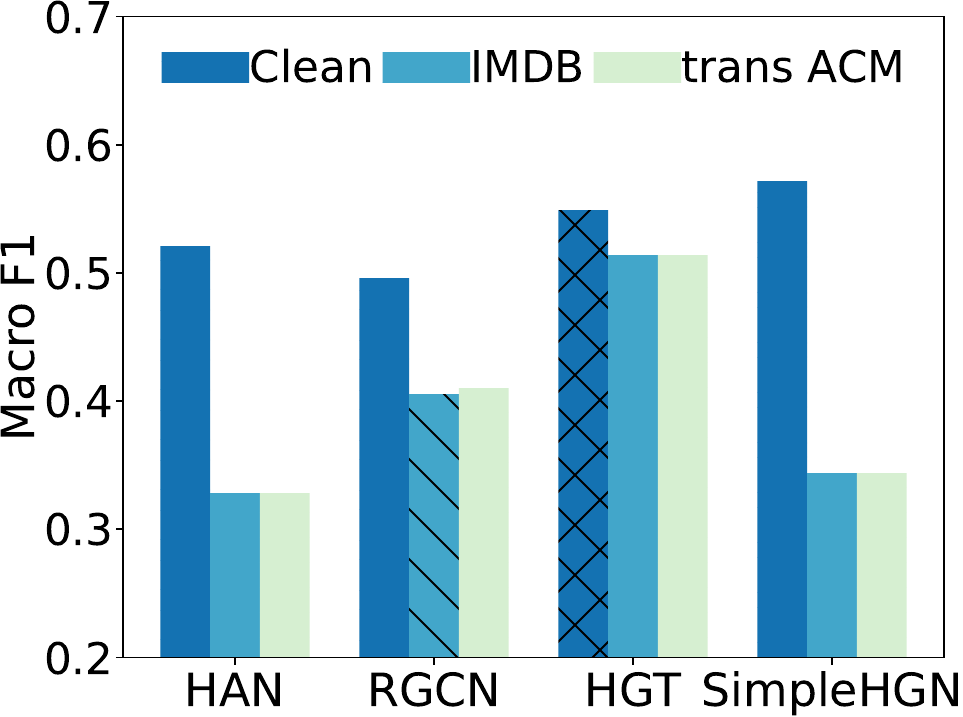}
		\caption{HeTa (1\% injection rate)}
		\label{MY}
	\end{subfigure}
	\centering
	\begin{subfigure}{0.24\linewidth}
		\centering
		\includegraphics[width=0.9\linewidth]{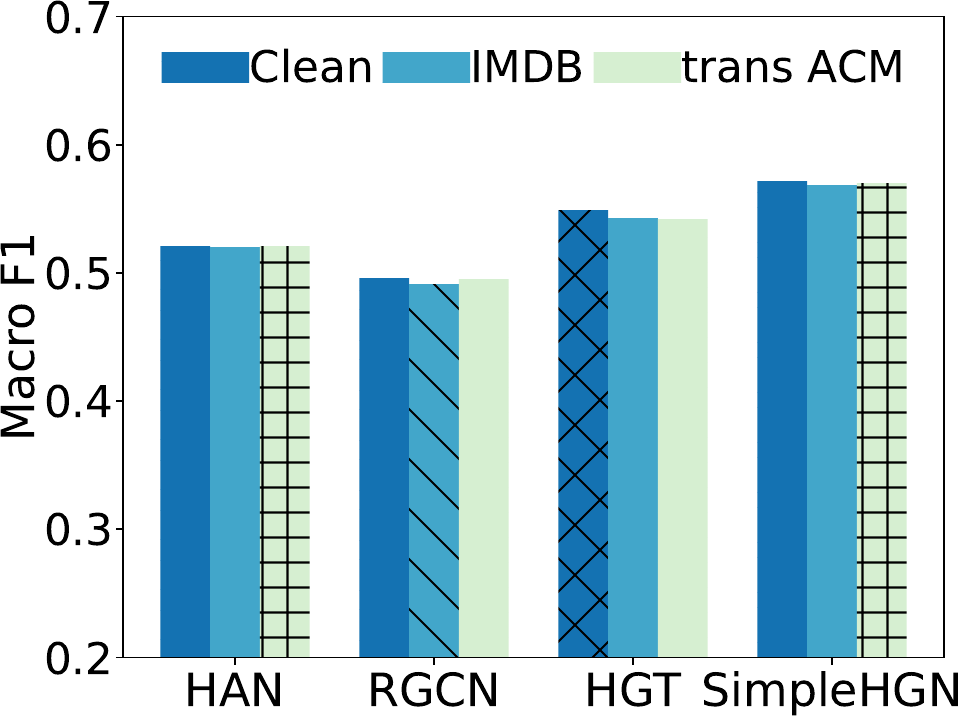}
		\caption{\(\text{G}^2\text{A2C} \) (1\% injection rate)}
		\label{G2A2C}
	\end{subfigure}
	\centering
    \centering
	\begin{subfigure}{0.24\linewidth}
		\centering
		\includegraphics[width=0.9\linewidth]{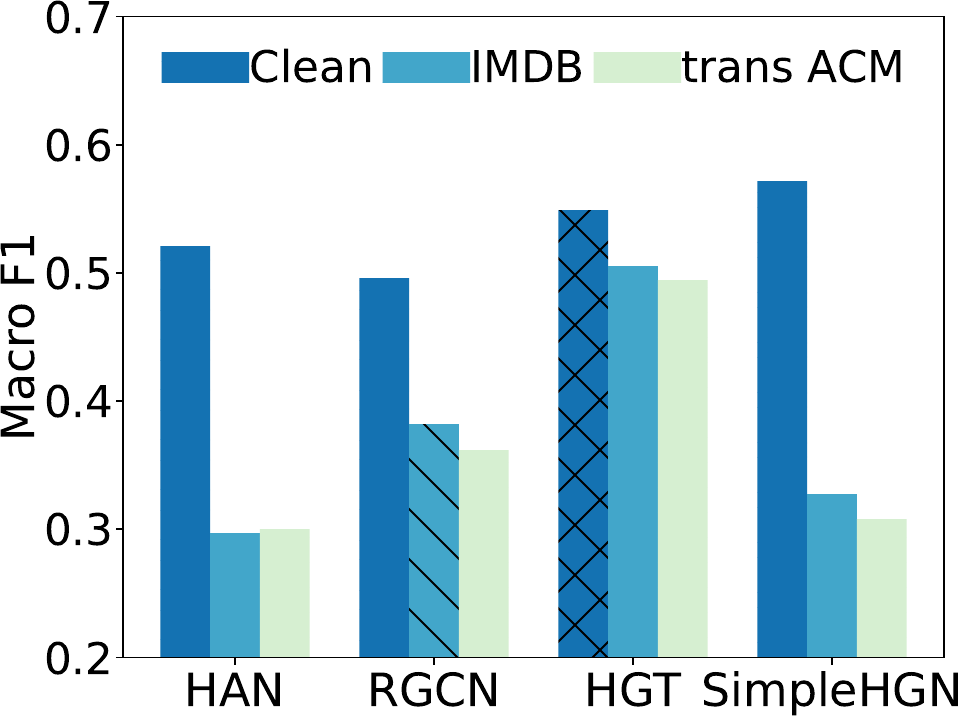}
		\caption{HeTa (2\% injection rate)}
		\label{MY}
	\end{subfigure}
	\centering
	\begin{subfigure}{0.24\linewidth}
		\centering
		\includegraphics[width=0.9\linewidth]{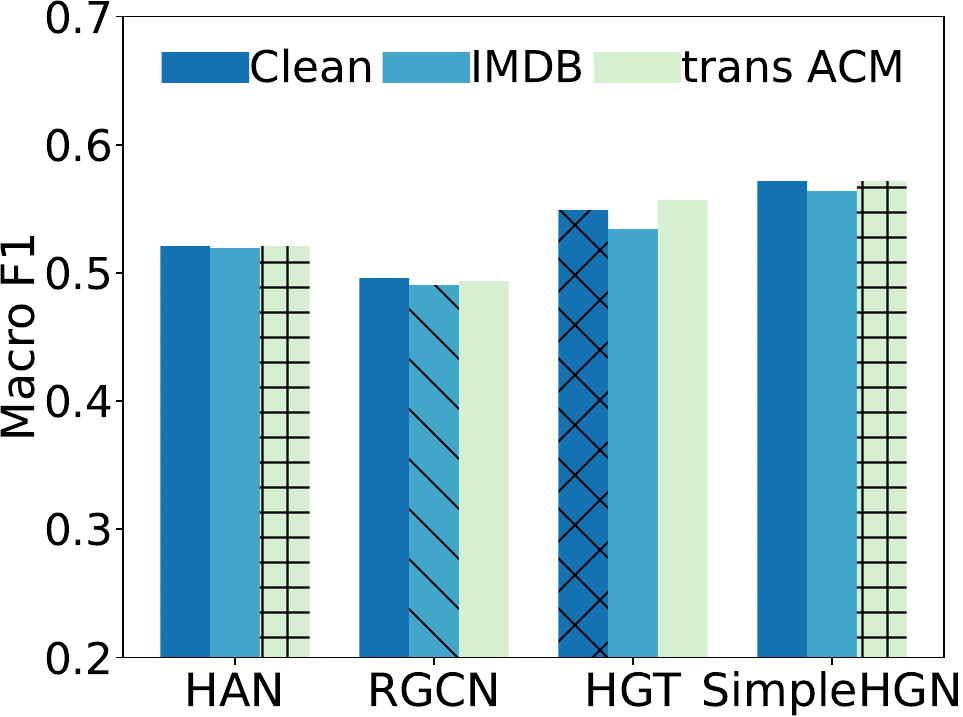}
		\caption{\(\text{G}^2\text{A2C} \) (2\% injection rate)}
		\label{G2A2C}
	\end{subfigure}
	\centering
    
	\caption{Supplementary material for Figure \ref{fig:transfer_data}. `IMDB’ means attack train on IMDB dataset. ‘transACM’ means parameters obtained from ACM dataset and applied to IMDB. }
	\label{fig:apptransfer_data}
\end{figure*}
\subsection{Implementation Details}\label{app:parameter_setting}
We implement HeTa based on Pytorch\footnote{\url{https://pytorch.org/}}. 
Each experiment is conducted five times, and the average values are reported. In attacks, lower scores indicate better attacking ability. To compare the effects of different attacks, we set various injection ratios \(\rho\), specifically 1\%, 2\%, and 5\%.
The penalty term \(\beta\) is set to 1.8. For the DBLP, ACM, and IMDB datasets, the scaling factor \(\alpha\) in Eq.\eqref{eq:Z_softmax} is [0.1,0.3,0.3].
For the DBLP, ACM, and IMDB datasets, hidden dimension is [128,64,256], and the number of hidden layers is [3,2,2], and the temperature on softmax is [0.4,0.5,0.8], and learning rate is [1e-2,1e-4,1e-3] with Adam optimizer, and K is [5,5,4], and the scaling factor \(\alpha\) in Eq.\eqref{eq:Z_softmax} is [0.1,0.3,0.3]. The confidential level \(k\) in Eq.\eqref{eq:loss_cw} is set to 0, control factor $r$ in Eq.\eqref{eq:loss_v} is set to 4. The penalty term \(\beta\) is set to 1.8.
Our fake node generator, have 3 layer on three datasets. For the DBLP, ACM, and IMDB datasets, the hidden dim is [64,128,128], learning rate is [1,1e-3,1e-2]. In addition, we implement early stopping with a patience of 10 epochs to prevent overfitting. To ensure fair comparisons, we fine-tune the parameters of various baselines using a grid search strategy. In \(\text{G}^2\text{A2}\text{C} \), for the DBLP, ACM, and IMDB datasets, hidden dim is [1024,32,512], feature budget is [1.5,1,1.5], node budget is [6,6,6].

We run all the baselines and carefully tuned the parameters to achieve the best performance. The implementation details of the baselines are as follows.
For FGA, we randomly select a certain number of target nodes in test set, with the quantity matching the number of nodes we injected, the budget for each target node is set to 5. We regard the nodes as  injected node when it connected to the target node after attack,  with the assumption that their features remain unchanged.
For RoHe-attack, we adopt the same setup as FGA, randomly selecting target nodes in the test set, with a perturbation budget of 5 for each target node, and other settings are based on the paper's specifications.
For \(\text{G}^2\text{A2}\text{C} \), we adopt the same random selection strategy and extend it to a global attack.
\section{Training Algorithm} \label{app:algorithm}
The pseudo code of HeTa is provided in Algorithm \ref{alg:algorithm}.

\begin{figure*}[h]
	\centering
	\begin{subfigure}{0.24\linewidth}
		\centering
		\includegraphics[width=0.9\linewidth]{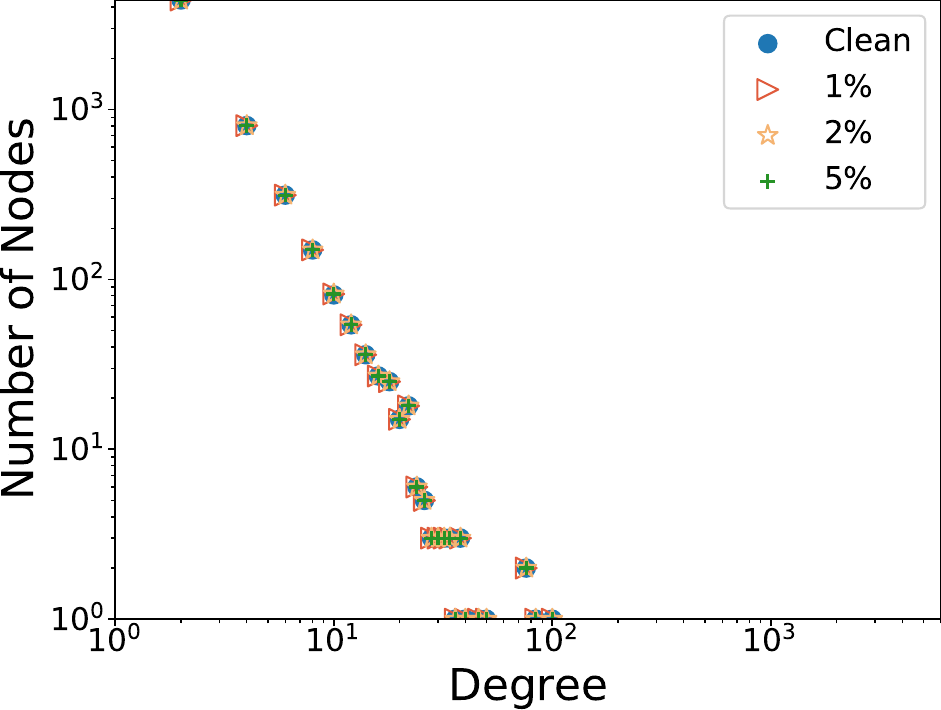}
		\caption{}
		\label{HAN}
	\end{subfigure}
	\centering
	\centering
	\begin{subfigure}{0.24\linewidth}
		\centering
		\includegraphics[width=0.9\linewidth]{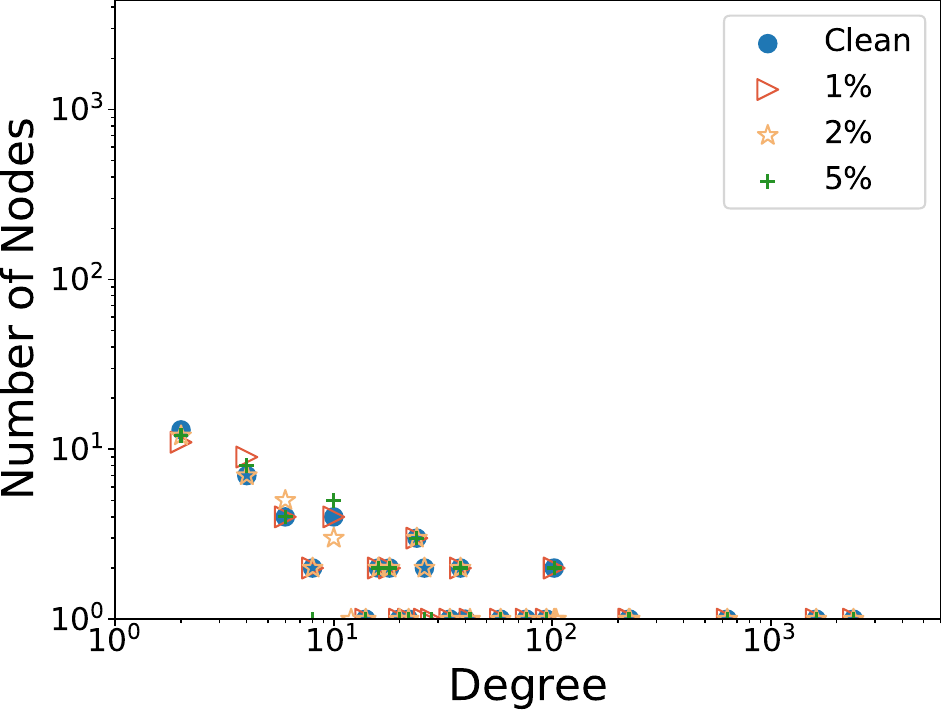}
		\caption{}
		\label{SimpleHGN}
	\end{subfigure}
        \begin{subfigure}{0.24\linewidth}
		\centering
		\includegraphics[width=0.9\linewidth]{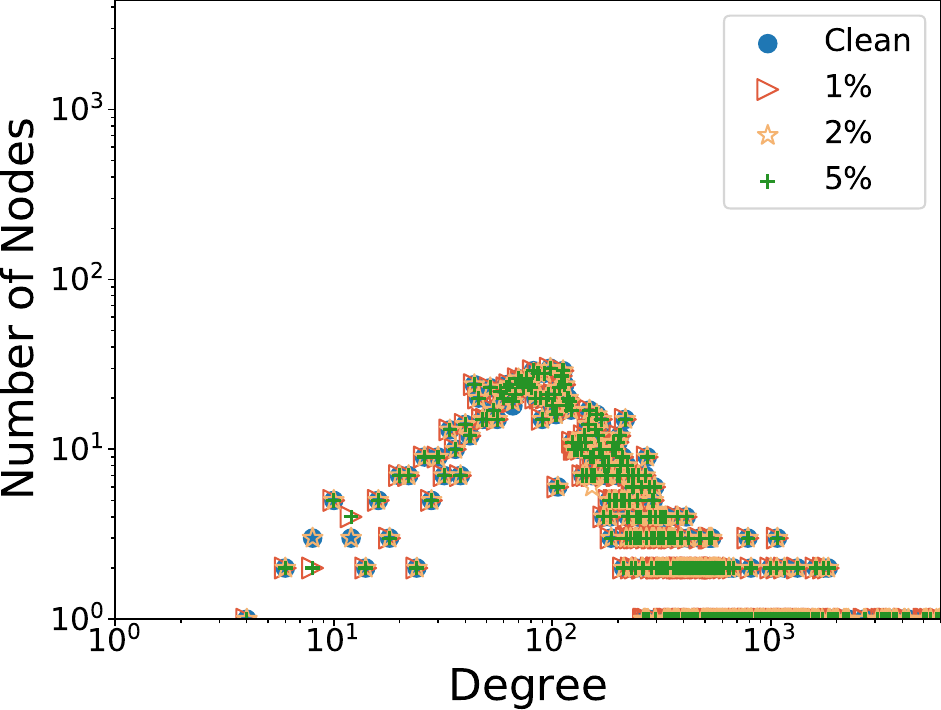}
		\caption{}
		\label{RGCN}
	\end{subfigure}
        \begin{subfigure}{0.24\linewidth}
		\centering
		\includegraphics[width=0.9\linewidth]{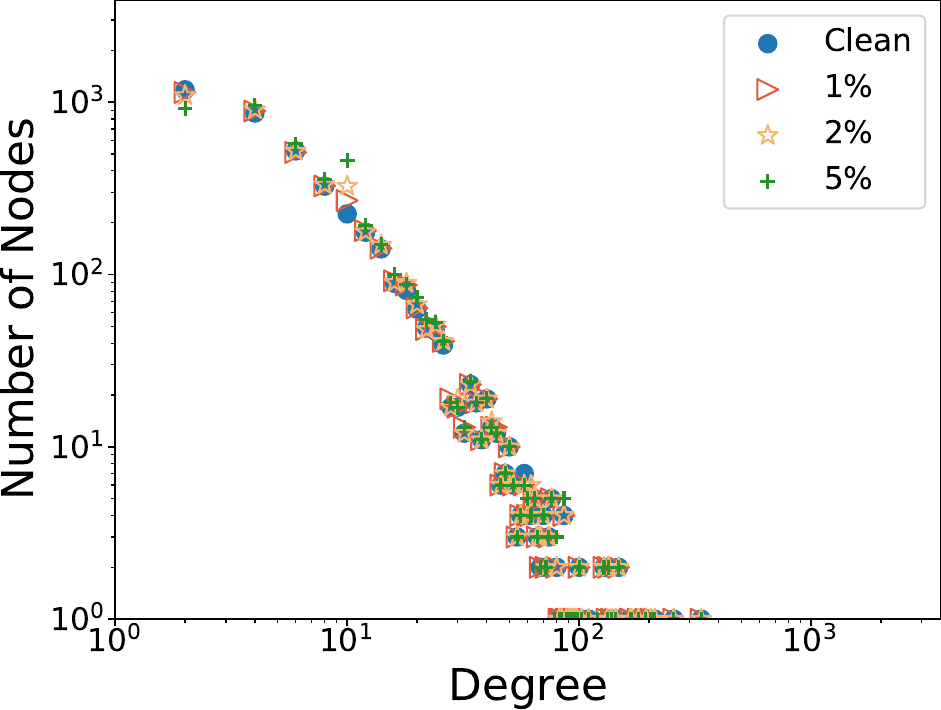}
		\caption{}
		\label{HAN}
	\end{subfigure}
	\centering
	\centering
	\begin{subfigure}{0.24\linewidth}
		\centering
		\includegraphics[width=0.9\linewidth]{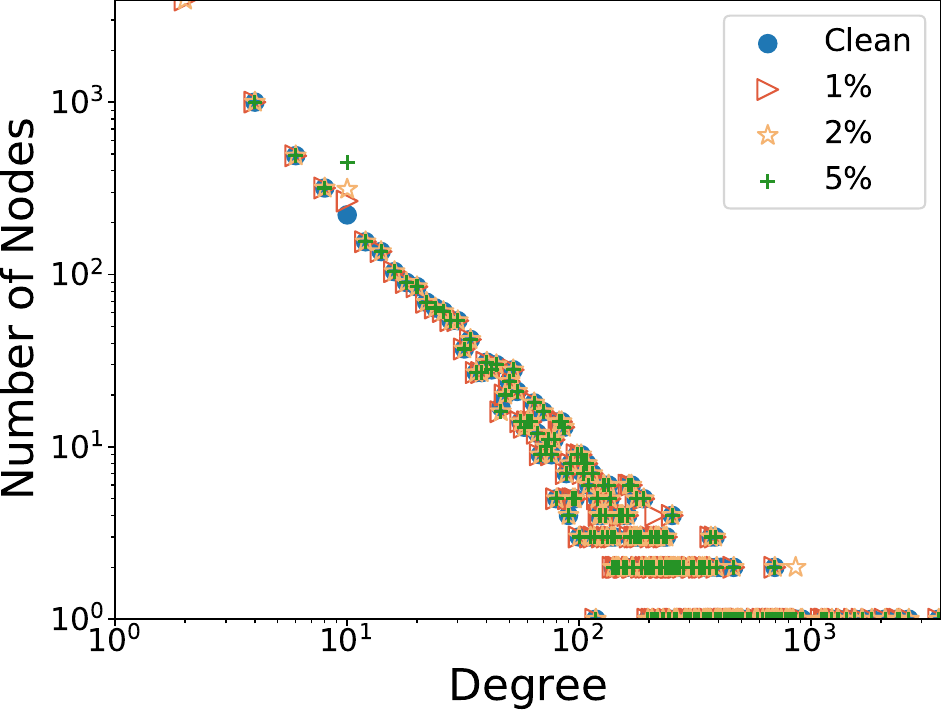}
		\caption{}
		\label{SimpleHGN}
	\end{subfigure}
    \begin{subfigure}{0.24\linewidth}
		\centering
		\includegraphics[width=0.9\linewidth]{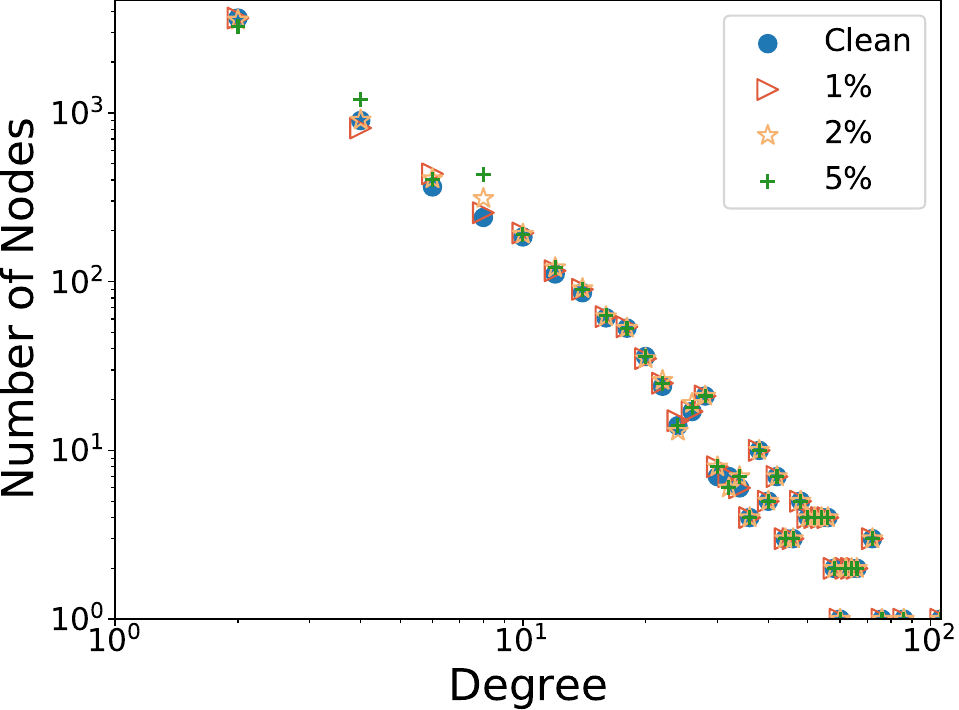}
		\caption{}
		\label{HAN}
	\end{subfigure}
    \begin{subfigure}{0.24\linewidth}
		\centering
		\includegraphics[width=0.9\linewidth]{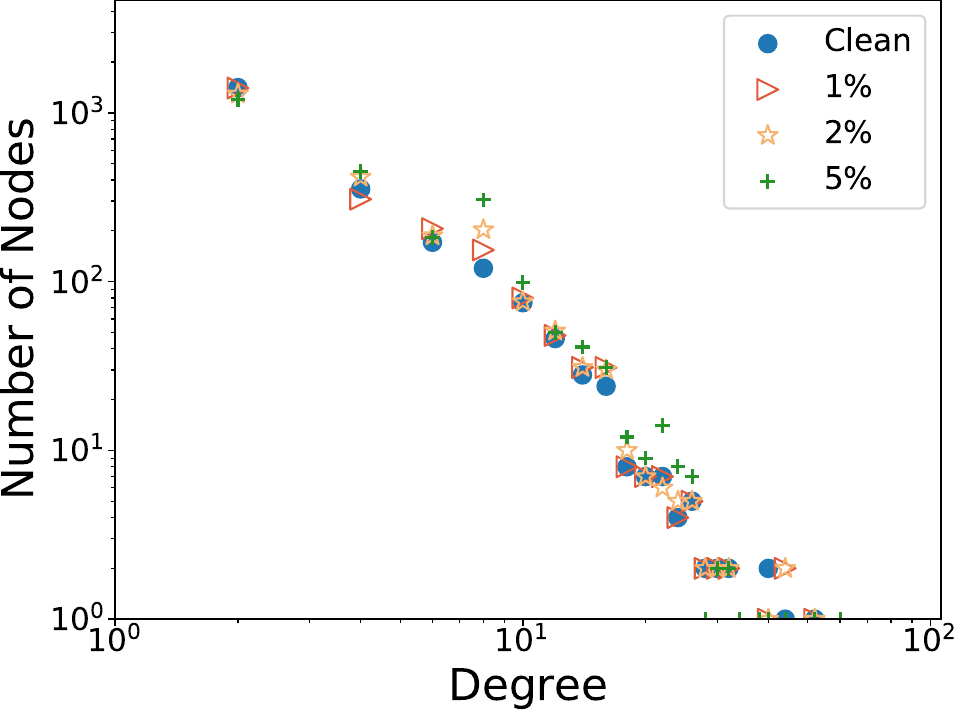}
		\caption{}
		\label{HAN}
	\end{subfigure}
    
	\caption{Node degree distribution under different injection rates.}
	\label{fig:degree}
\end{figure*}

\section{More Experiment Results}
\subsection{More Hyper-parameter Study} \label{app:Hyper-parameter_appendix}
We conduct more hyper-parameter experiments on the degree of injected nodes \(K\) and the update rule \(\beta\) in Figure \ref{parameter_both_appendix}. 

\subsection{Attack effectiveness}
\textbf{In our relation-wise setting, target nodes with lower degrees are more susceptible to attacks.} The simplified version of HeTa (\textit{Simple}) involves selecting nodes with lower degrees as neighbors for the injected nodes, guided by the relation weight  $\boldsymbol{\mu}$ as shown in Table \ref{tab:Simple_Model_Efficiency}. 
By comparing \textit{Simple} with HeTa and the RoHe-attack, \textit{Simple} achieves ranking among the top two in overall performance and the DBLP dataset are more vulnerable under the simplified version of the attack. 
This is likely related to the degree distribution~\cite{zhang2024maximizing,zou2021tdgia}, as DBLP has a high proportion of nodes with low degrees.
\paragraph{Effective Destruction of Label Distribution with Low Injection Rate.} To visually demonstrate the impact of the attack model on the target model's label distribution, we observe the changes in the model's classification labels under different adversarial injection rates, as shown in Figure~\ref{fig:label_distribution}. With the increase in injection rates, the proportion of labels changes significantly, especially in the ACM dataset. 
Indicates that our attack method remains effective even when the number of adversarial nodes is limited.


\paragraph{Unnoticeable Perturbation on Degree Distribution.}
We examine the graph structure to assess  if the degree distribution shifts after the attack, the findings shown in Figure~\ref{fig:degree}. The similarity between the degree distributions of the original and the adversarial graphs suggests that the perturbation edges have been effectively integrated without drawing attention to specific nodes. Since HeTa selects nodes (e.g., low-degree nodes) to connect with a lower budget guided by the relation weights. This approach minimizes the risk of detection and maintains the overall integrity of the graph structure. 

\paragraph{About the Parameter Size and Time Complexity.}
Our surrogate model consists of three components: the Projector, Message Passing, and Classifier, totaling $\mathcal{O}\left(2 d^{2}+K|\mathcal{R}|\right)$ parameters. In comparison, as a widely used HGNN, HAN’s parameter size is primarily driven by its multi-head attention, resulting in a total of $\mathcal{O}\left(
2K\left(|\mathcal{R}| N +1 \right)d^{2} + 2 Kd
\right)$. Here, $K$ is HGNN layers, $d$ is the embedding dimension, $|\mathcal{R}|$ and $N$ are the number of relations and attention heads, respectively.
This shows that our surrogate model is more lightweight. The time complexity is primarily driven by the Fake Node Generator and Fake Edge Selector. 
The complexity for node feature updates $O(3d^{2} )$, where $d$ is the embedding dimension. The complexity for the fake edge selector is
$O(n)$, due to its simplified calculation using vectors and $n$ is total nodes. Therefore, the overall time complexity for injecting $H$ fake nodes after $M$ time step is $O(HM(3d^{2}+n))$.

\paragraph{Model Efficiency and Scalability.}
We test the model's cost on datasets of varying sizes, including a larger graph (Freebase), as shown in Table 1. 1) Efficiency: The runtime increases with dataset size, but our model remains highly efficient. e.g., on the Freebase dataset, the model trains in just 3.77 minutes. 2) Scalability: HeTa effectively attacks larger graphs, e.g., Freebase. We test it with a 0.01 node injection on RGCN, and the F1 score dropped from 0.5774 on clean data to 0.2205 after the attack, a reduction of \textbf{up to 60.44\%}. This efficiency and scalability are due to HeTa’s tailored design, including its lightweight and strategies such as gradient simplification (Eq. 11).

\begin{table} [H] 
\small
\centering
\caption{Cost comparison of different dataset sizes: Runtime (Minutes), Memory (GB), and Parameters (K).}
\begin{tabular}{cccccc}
\toprule
Dataset & \#Nodes & \#Edges &  Runtime & Memory& Parameter \\
\midrule
IMDB&12772& 37288& 0.99m&1.3G & 1706K\\
ACM&10942& 547872& 1.21m &1.55G &1616K\\
Freebase&43954& 151034 & 3.77m & 6.13G & 8684K\\
\bottomrule
  \end{tabular}
   \label{cost}
\end{table}

\end{document}